\documentclass[10pt,twocolumn,letterpaper]{article}

\usepackage{cvpr}              
\usepackage{multirow}
\usepackage[accsupp]{axessibility}  

\usepackage[dvipsnames]{xcolor}

\newcommand{\dst}{dynamic tokens\xspace}

\usepackage{tcolorbox}
\tcbuselibrary{skins,breakable}
\newtcolorbox{commandbox}{
  enhanced,
  breakable,
  colback=blue!5,
  colframe=blue!40!black,
  rounded corners,
  boxrule=0.7pt,
  left=6pt,
  right=6pt,
  top=6pt,
  bottom=6pt,
  fontupper=\normalsize, 
  sharp corners=northwest,
  before skip=10pt,
  after skip=10pt
}

\definecolor{cvprblue}{rgb}{0.21,0.49,0.74}
\usepackage[pagebackref,breaklinks,colorlinks,citecolor=cvprblue]{hyperref}

\usepackage{hyperref}
\usepackage{url}
\usepackage{comment}
\usepackage{booktabs}
\usepackage{graphicx}   
\usepackage{tikz}
\usepackage[table]{xcolor}
\usepackage{lipsum} 
\usepackage{skak}

\definecolor{tabfirst}{rgb}{1, 0.7, 0.7} 
\definecolor{tabsecond}{rgb}{1, 0.85, 0.7} 
\definecolor{tabthird}{rgb}{1, 1, 0.7} 

\def\paperID{6107} 
\def\confName{CVPR}
\def\confYear{2026}

\title{Velox: Learning Representations of 4D Geometry and Appearance}

\author{
Anagh Malik$^{\symknight\symrook}$ \;
Dorian Chan$^\symknight$ \; Xiaoming Zhao$^\symknight$ \; David B. Lindell$^\symrook$ \; \\ Oncel Tuzel$^\symknight$ \; Jen-Hao Rick Chang$^\symknight$ \\ \\
$^\symknight$Apple  \;\; $^\symrook$University of Toronto \\
    \small{\url{https://apple.github.io/ml-velox}}
}

\begin{document}

\twocolumn[{%
\renewcommand\twocolumn[1][]{#1}%
\maketitle
\begin{center}
    \captionsetup{type=figure}
        \includegraphics[width=\textwidth, height=0.5\textwidth]{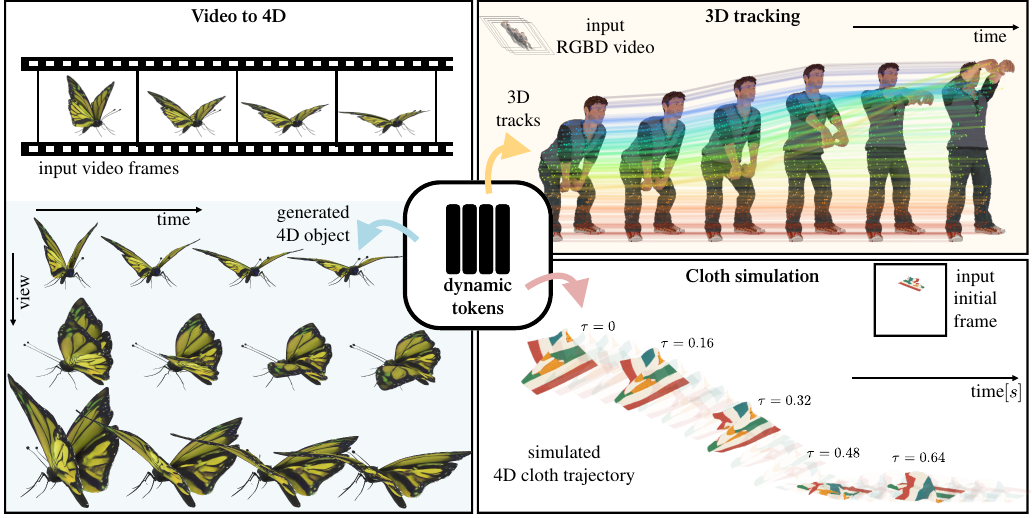}
    \\
    \captionof{figure}{\textbf{Velox} learns a latent \textbf{4D object representation}, which we call \textit{\dst}. This representation proves useful across diverse downstream tasks. 
\textbf{Left:} Generating directly in our latent space enables \textit{video-to-4D generation}. 
\textbf{Top right:} A network trained atop our representation can \textit{track 3D points} in input RGBD videos.
\textbf{Bottom right:} Via an image-to-4D pipeline, \dst can also be used for \textit{cloth simulation} given an initial input frame. Asset credits~\cite{dylansburner,kettlebell_swing}.}
    \label{fig:teaser}
\end{center}
}]


\maketitle
\begin{abstract}

We introduce a framework for learning latent representations of 4D objects which are descriptive, faithfully capturing object geometry and appearance; compressive, aiding in downstream efficiency; and accessible, requiring minimal input, \ie, an unstructured dynamic point cloud, to construct. Specifically, Velox trains an encoder to compress spatiotemporal color point clouds into a set of \textit{\dst}. These tokens are supervised using two complementary decoders: a 4D surface decoder, which models the time-varying surface distribution capturing the geometry; and a Gaussian decoder, which maps the tokens to 3D Gaussians, helping learn appearance.
To demonstrate the utility of our representation, we evaluate it across three downstream tasks---video-to-4D generation, 3D tracking, and cloth simulation via image-to-4D generation---and observe strong performances in all settings.
Please see the \href{https://apple.github.io/ml-velox}{website} for video results.
\vspace{-2em}
\end{abstract}    
\section{Introduction}

Progress in 2D and 3D vision has highlighted the central role of representation learning. 
Self-supervised methods~\cite{caron2021emerging, oquab2024dinov2, simeoni2025dinov3, chen2020simple, he2020momentum, grill2020bootstrap, caron2020unsupervised, zbontar2021barlow} demonstrate that learning to reconstruct or structure data produces features that transfer effectively to downstream tasks.
Latent representations have also become fundamental to generation and manipulation~\cite{rombach2021highresolution, xiang2024structured}.
Motivated by these insights, we investigate whether similar principles extend to the spatiotemporal domain.
In particular, we explore whether a 4D representation of dynamic objects learned purely from reconstruction can provide a versatile foundation for diverse applications.
An effective 4D representation should be \textit{descriptive}, faithfully capturing scene geometry and appearance over time; \textit{compressive}, maintaining a compact form that facilitates downstream processing; and \textit{accessible}, relying only on readily obtainable inputs. 
Unfortunately, existing methods do not satisfy all of these requirements. 
Many are tailored to specific tasks such as view synthesis~\cite{keil2023kplanes,liu2023robust,li2022neural3dvideosynthesis,pumarola2020d,fang2022tineuvox,shao2023tensor4d,wang2022mixed,wu20244dgaussian,luiten2023dynamic,yang2023deformable3dgs}, character animation~\cite{tan2018variational, kim2024meshupmultitargetmeshdeformation}, or 3D tracking~\cite{xiao2024spatialtracker, xiao2025spatialtrackerv2}, and thus may discard information critical for general scene understanding. 
Moreover, they often leverage high-dimensional representations that are computationally unsuitable for disparate downstream tasks. 
Conversely, efforts toward generalizable 4D object representations often omit appearance modeling~\cite{rempe2020caspr,wei2024motion,zhang2024dnf,erkoç2023hyperdiffusion}, limiting their descriptiveness; or rely on specialized encoder inputs, \eg, temporal correspondences~\cite{zhang2025gaussian,yenphraphai2025shapegen}, making them inaccessible for downstream tasks.

To tackle these limitations, we propose \textbf{Velox}: a method to learn a faithful representation of 4D scene geometry and appearance. 
It compresses input 4D point clouds into \textit{\dst}, achieving more than $30\times$ compression. 
Utilizing a Perceiver-style encoder~\cite{perceiverio, zhang20233dshape2vecset}, Velox aggregates information across space and time without the need for correspondences, making it easily accessible for downstream tasks. 
To train this encoder, we utilize two complementary decoders: a 4D surface decoder, which relies on flow-matching to model precise geometry as a function of time~\cite{chang20243dshapetokenization}; and a 3D Gaussian decoder~\cite{chang2026lito,xiang2024structured}, which helps learn object appearance. 
We train the representation on a curated dynamic object dataset~\cite{deitke2022objaverse,deitke2023objaversexl}, which we augment with additional textures to improve diversity.

We conduct a comprehensive evaluation of the proposed method across four complementary settings and show state-of-the-art results.
First, we assess reconstruction quality against baselines that either encode scenes independently per timestep or rely on explicit temporal correspondences.
Secondly, we train a generative model that maps monocular video to \dst, showcasing high-quality geometry and appearance generation. 
%
Third, we train a 3D RGBD tracking model directly on the learned \dst, demonstrating that our approach can produce precise 3D trajectories.
Finally, we adopt \dst as the underlying representation for cloth simulation, illustrating its flexibility in modeling dynamic, deformable objects.
In summary, we make the following contributions: 
\begin{enumerate}
    \item A framework for learning representations that jointly and effectively model 4D geometry and appearance without relying on correspondence as encoder input.
    \item A comprehensive evaluation of the representation’s reconstructive ability compared with per–time-step and deformation-based methods.
    \item Extensive demonstrations showing that \dst are effective across various tasks, including video-to-4D generation, 3D tracking, and cloth simulation, consistently achieving performance competitive with SOTA methods.
\end{enumerate}

\begin{figure*}[htbp]
    \centering
\includegraphics[width=\textwidth]{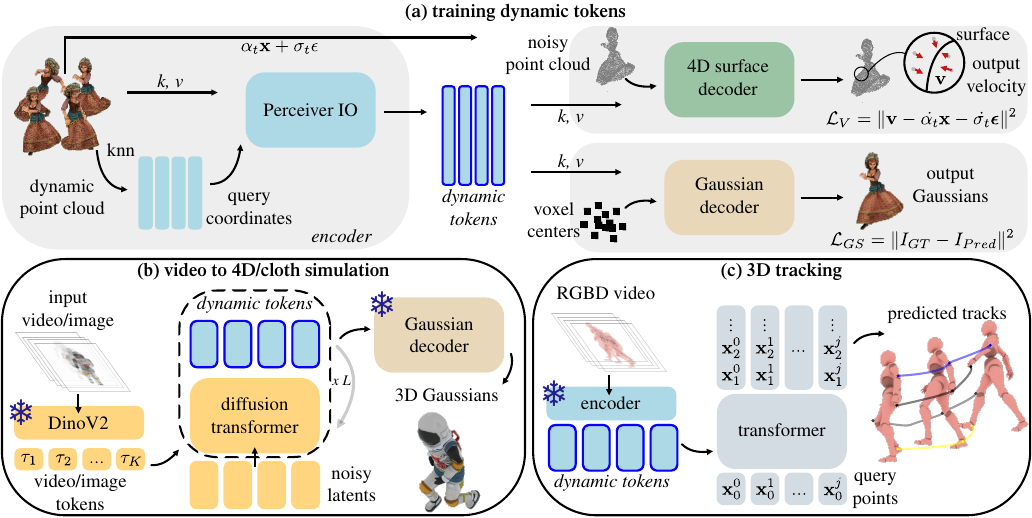}
    \caption{\textbf{Method.} \textbf{(a)} We use a Perceiver IO encoder~\cite{perceiverio, zhang20233dshape2vecset} to map dynamic point clouds and queries to \dst. We train this representation with two decoders: the 4D surface decoder, which maps noisy input surface points to denoised points; and a Gaussian decoder, which maps from voxel centers to 3D Gaussian~\cite{kerbl3Dgaussians} parameters. \textbf{(b)} To train our models for video-to-4D and cloth simulation, we use a DiT~\cite{dit} that directly generates 4D representations in the form of \dst conditioned on input image/video DinoV2 features. We can then use the Gaussian decoder to map the \dst to 3D Gaussians. \textbf{(c)} Conditioned on \dst calculated from an input RGBD video, we train a tracking network that maps 3D query points in the initial frame $\mathbf{x}_0^i$ to their future 3D locations $\mathbf{x}_j^i$. Asset credit:~\cite{astro13_run,peasant_girl_breakdance_1990,Mirandanimator_CharacterXBot_2014}.}
    \label{fig:method}
        \vspace{-1em}
\end{figure*}

\section{Related Work}
\label{sec:related_work}

Our work focuses on 4D latent representation learning, with downstream applications in 3D tracking and object generation.
In this section, we review methods directly related to 4D latent representations; a summary of their key properties is provided in the appendix.
Additional related work on 2D/3D tracking~\cite{karaev2024cotracker, sand2008particle, karaev2024cotracker3, harley2022particle, le2024dense, wang2023tracking, xiao2024spatialtracker, wang2024scenetracker, ngo2024delta, zhang2025tapip3d, jin2024stereo4d, feng2025startrack, wang2024dust3r}, 4D  generation~\cite{jiang2024consistentd, bah20244dfy, ren2023dreamgaussian4d, singer2023text4d, bah2024tc4d, gao2024gaussianflow, ling2024align, yin20234dgen, zhao2023animate124, zheng2024unified, zeng2024stag4d, wu2024cat4d, xie2024sv4d, yao2024sv4d2, pan2024efficient4d, liang2024diffusion4d, ren2024l4gm, zhang2025gaussian, yenphraphai2025shapegen}, and deformable-object simulation~\cite{zhang2024adaptigraph, huang2022mesh, helearning, shi2024robocraft, shi2023robocook, ai2024robopack, tian2025uniclothdiff} is also deferred to the appendix.

4D representation learning is a relatively recent direction compared to the extensive literature on static 3D representations~\cite{zhu2025ponderv2paveway3d, yang2019pointflow, wu2022ptv2, wu2024ppt, pointcept2023, zhang20233dshape2vecset, zhao2024michelangelo, wu2024direct3d, zhang2024clay, vahdat2022lion, luo2021diffusion, chang2026lito, chen2025dora, xiang2024structured, hunyuan3d22025tencent}.
While in principle a dynamic object can be represented by concatenating independent 3D representations at each time instance, this ignores temporal continuity and typically yields high-dimensional, temporally inconsistent features, often manifesting as jitter or unstable geometry.

An alternative line of work learns representations directly from video~\cite{lal2021coconets, herley2019embodied, wang2025cut3r, carreira2024scaling4drepresentations} by maintaining a latent state over time. 
However, such representations are constrained by video visibility --- since only surfaces observed in the input frames are available, they generally lack complete object geometry and cannot recover full shapes.

Closely related to our setting are object-centric 4D representations that aim to model complete spatiotemporal shape information~\cite{erkoç2023hyperdiffusion, zhang2024dnf, wei2024motion, rempe2020caspr, zhang2025gaussian, yenphraphai2025shapegen}. 
One line of work~\cite{erkoç2023hyperdiffusion, zhang2024dnf} treats neural-field weights as the representation, but requires optimizing a separate neural field for each object, limiting scalability.
Many other approaches rely on temporal correspondences between 3D points as input to the encoder or as explicit supervision during training~\cite{wei2024motion, rempe2020caspr, zhang2025gaussian, yenphraphai2025shapegen}. 
Such correspondences define deformations across time, but are challenging to obtain for complex dynamic scenes, which restricts usable training data and constrains inference pipelines of their encoders.
%
%
Additionally, these methods typically only model geometry~\cite{wei2024motion, rempe2020caspr, yenphraphai2025shapegen} or separate geometry and appearance into distinct representations~\cite{zhang2025gaussian}. 

In contrast, our method requires no correspondences and learns a single latent representation that jointly captures dynamic geometry and appearance, enabling a simpler training pipeline and broader applicability, \eg, 3D tracking.

%


\section{Method}
\label{sec: method}

We learn a 4D object representation in the form of \textit{\dst} $\textbf{s}$, which contains a set of $k$ tokens of dimension $d$. 
Specifically, an encoder $E$ takes as input an unstructured spatiotemporal color point cloud:
\begin{align}
    \mathcal{X}
 = \{(\textbf{x}_i \in \mathbb{R}^3, \textbf{c}_i \in [0, 1]^3, \tau_i \in \mathbb{R})\}_{i=1}^{N}, \label{eq:input}
\end{align}
where $\textbf{x}_i$, $\textbf{c}_i$ and $\tau_i$ denote the spatial location, RGB color and time of a point $i$, respectively. 
To ensure our representation $\textbf{s} = E(\mathcal{X})$ captures time-varying geometry and appearance, we jointly supervise it to model 4D geometry and appearance.
The entire model can be trained only with multiview RGBD videos (and their backprojected point clouds), which can be easily rendered from any animated scene, unlike Signed Distance Function (SDF)/occupancy representations that require watertight meshes and significant preprocessing.
In the rest of this section, we provide a high-level description of our encoder and decoders. Please refer to the supplement for more details on our exact architectures. See Figure~\ref{fig:method} for an overview.

\vspace{-1em}
\paragraph{Spatiotemporal patch encoder.} 
Inspired by recent work in 3D representations~\cite{chang2026lito,zhang20233dshape2vecset}, we aim to encode patches of the 4D spatiotemporal surface, \ie, local time-varying geometry/appearance, into \dst. 
To do this, we leverage a Perceiver IO~\cite{perceiverio} architecture as our encoder, with $k$ points sampled from the input point cloud $\mathcal{X}$ as queries. These queries cross-attend only to the points in $\mathcal{X}$ that are closest to it, \ie, each query only observes its local neighborhood in the 4D space. 
Windowed self-attention layers propagate this local patch information to other queries while maintaining computational efficiency.
Our final latents are of dimension $8192\times 32$, a similar order-of-magnitude to recent work in 3D~\cite{chang2026lito,xiang2024structured} and 4D~\cite{zhang2025gaussian}.

\vspace{-1em}
\paragraph{Learning geometry.} 
We design a 4D surface decoder inspired by~\citet{chang20243dshapetokenization}, which employs a flow-matching decoder that models 3D surfaces as a probabilistic density function in the 3D space, $p(\mathbf{x} \vert \mathbf{s})$. 
A natural extension would be to model the 4D surface as a joint distribution $p(\mathbf{x}, \tau \vert \mathbf{s})$. 
However, such a formulation makes it difficult to directly sample the points corresponding to a particular target time---a potentially useful task for downstream applications, \eg, rendering the discrete frames of a video of a dynamic point cloud. 
Thus, for a given scene, we instead model the conditional distribution \( p(\mathbf{x} \vert \tau, \mathbf{s}) \), representing the probability of a 3D point \( \mathbf{x} \) given a specific time \( \tau \). 
Assuming uniformly distributed $p(\tau)$, the conditional distribution is simply a scaled version of the joint distribution. 

Formally, the 4D surface decoder \( V \) takes as input the latent \( \mathbf{s} \), the flow-matching time \( t \), the target frame time \( \tau \), and a noisy surface point \( \mathbf{x}_t = \alpha_t\mathbf{x} + \sigma_t\boldsymbol{\epsilon} \), where $\mathbf{x}$ is randomly selected from points in $\mathcal{X}$ with the same frame time $\tau$, and \( \boldsymbol{\epsilon} \) is drawn from the standard normal. The decoder then predicts the flow-matching velocity at that location \( \mathbf{v} = V(\mathbf{s}, \mathbf{x}_t, t, \tau) \), which we supervise with the loss:
\begin{equation}
\mathcal{L}_V = \lVert \mathbf{v} - \dot{\alpha_t}\mathbf{x} - \dot{\sigma_t}\boldsymbol{\epsilon} \rVert^2.
\label{eq: geometry loss}
\end{equation}
A point cloud corresponding to the surface at time $\tau$ can then be sampled by integrating $\mathbf{v}$ from $t=0$ to $t=1$. To implement this, we use the architecture of Chang et al.~\cite{chang20243dshapetokenization} with an additional positionally encoded input $\tau$.

\vspace{-1em}
\paragraph{Learning appearance.}
To supervise the appearance encoded by \dst---and allow images to be rendered from them for downstream tasks---we decode 3D Gaussians~\cite{kerbl3Dgaussians}.
An ideal decoder would decode all Gaussians for the entire duration of the scene simultaneously, but such an approach is extremely memory intensive. 
A more efficient approach would be to predict a canonical set of 3D Gaussians and their deformations over time~\cite{luiten2023dynamic,zhang2025gaussian}, but such deformations are ill-defined for scenes with appearing or disappearing objects.
Instead, we train a model $G(\mathbf{s}, \tau)$, which directly maps \dst to 3D Gaussians for an input timestamp $\tau$. 
In practice, to aid with convergence, like recent work in 3D representations~\cite{xiang2024structured,chang2026lito}, we decode Gaussians at only sparsely occupied voxels `$\text{Vox}$', using a Perceiver IO~\cite{perceiverio} architecture. 
Our supervision loss is thus given by:
\begin{equation}
\mathcal{L}_{GS} = \lVert I_{GT} - \text{Render}(G(\text{Vox}, \mathbf{s}, \tau), \text{H}_I) \rVert^2, \label{eq:gs_loss}
\end{equation}
where \(I_{GT}\) is the ground-truth image and \(\text{H}_I\) are the camera intrinsics and extrinsics. 

For `$\text{Vox}$', during training, we use ground-truth voxels calculated from the input point cloud. 
At inference, one can directly query the 4D surface decoder to construct `$\text{Vox}$'.
To avoid the time for sampling the point cloud, we train a downstream decoder that predicts occupied voxels from \dst, as the geometry information is embedded in \dst due to the geometry supervision \cref{eq: geometry loss}. 
This enables faster 3D Gaussian decoding during inference.
 
\vspace{-1em}
\paragraph{Training objective.} We supervise the \dst~$\mathbf{s}$ using both the 4D surface decoder loss for fine geometric supervision and the 3D Gaussian loss for appearance supervision. 
Additionally, we regularize the \dst~using a KL-divergence loss with the standard Normal distribution, which reduces to a simple \(L_2\) loss on the tokens,
$\mathcal{L}_C = \lVert \mathbf{s} \rVert^2.$ The final objective is given by:
\begin{equation}
    \mathcal{L} = \mathcal{L}_{V} + \mathcal{L}_{GS} + \gamma \mathcal{L}_{C}, 
    \end{equation}
where $\gamma = 10^{-4}$ in our experiments.

\subsection{Tasks}
To demonstrate the utility of our \dst, we tackle three tasks: video-to-4D generation, cloth simulation (via image-to-4D generation), and 3D tracking. We describe the specific methods used to achieve these tasks here (overview in \cref{fig:method}). See supplement for exact model details.

\vspace{-1em}
\paragraph{Video-to-4D/Cloth simulation.} For both the video-to-4D and cloth simulation applications, we aim to generate \dst~given a scene observation. 
We learn the video-to-4D generative model on dynamic objects in Objaverse, similar to prior works~\cite{ren2024l4gm,wu2024cat4d,xie2024sv4d,yao2024sv4d2}.
The conditioning video has 24 frames, with fixed and unknown camera parameters throughout the video.
For cloth simulation, we learn an image-to-4D generative model on a physics-simulated dataset containing randomly folded cloth falling from random heights~\cite{lips2024learning}, the conditioning image is the first frame of the simulated sequence. 
For both tasks, we use a DiT architecture~\cite{dit} with learnable zero-initialized positional encodings for the tokens. 
DINOv2~\cite{oquab2024dinov2} is used to encode the image(s).  
For video-to-4D, we provide the input time for each frame in the positional encoding.

\vspace{-1em}
\paragraph{3D tracking}
For 3D tracking, we unproject an input RGBD video and compute the \dst of the output point cloud.
We train a Perceiver IO encoder that takes as input query the 3D positions in the first frame and as key and value the \dst.
The model predicts the 3D positions of the query points at all future frames. 
We keep this setup simple, without any iterative refinement or fine-tuning~\cite{harley2022particle,karaev2024cotracker3}, to showcase the information encoded in \dst.

Many tracking methods also output visibility. For simplicity, we do not directly predict visibilities. 
Instead, we compare our estimated 3D track positions with the RGBD video. 3D tracks that are farther from the camera than the RGBD surface are marked as occluded.

\begin{figure}[tbp]
    \centering
\includegraphics[width=\columnwidth]{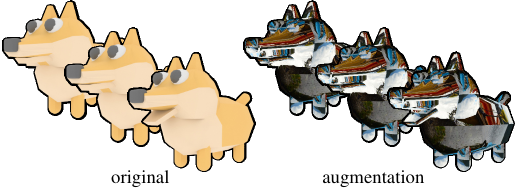}
\caption{\textbf{Texture augmentations.} We augment training sequences with random textures, increasing appearance diversity. Asset credit:~\cite{dog}.}
    \label{fig:augemntations}
    \vspace{-1em}
\end{figure}

\begin{figure*}[htbp]
    \centering
\includegraphics[width=\textwidth]{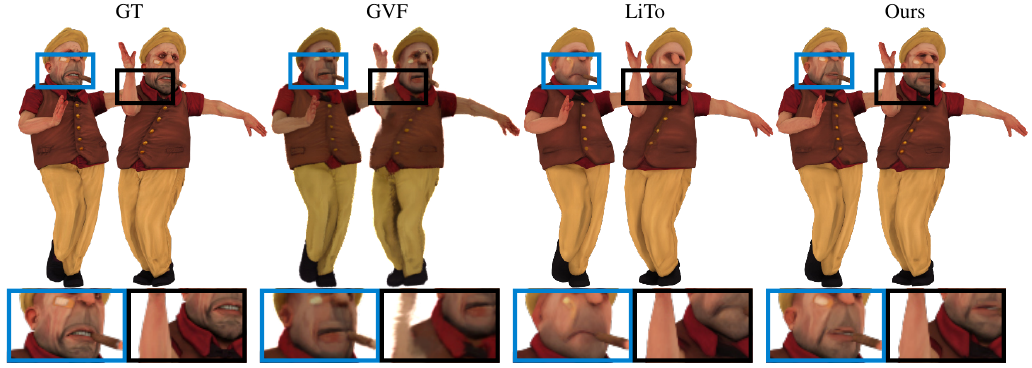}
    \vspace{-2em}
    \caption{\textbf{Reconstruction results.} \textbf{Black inset:} Our approach outperforms GVF~\cite{zhang2025gaussian}, especially on areas with bigger movements, where deforming Gaussians proves challenging. \textbf{Blue inset:} Furthermore, due to jointly modeling the object across time, we recover more faithful appearance than the single frame-based LiTo~\cite{chang2026lito}. Asset credit~\cite{salsa}.} 
    \label{fig:recon_res}
    \vspace{-1em}
\end{figure*}

\section{Dataset}

\paragraph{Dynamic objaverse}
We train all models (except those for cloth simulation) on dynamic objects  in Objaverse~\cite{deitke2022objaverse} filtered by L4GM~\cite{ren2024l4gm} and in Objaverse-XL~\cite{deitke2023objaversexl} filtered by Diffusion4D~\cite{liang2024diffusion4d}.
For each animation, we take the first two 24 frame sequences, as well as the first two 48 frame sequences skipping every other frame. 
This leaves us with about 110k total 24-frame animations, which we normalize to lie within scene bounds $[-1, 1]$. 

\vspace{-1em}
\paragraph{Texture augmentation}
We observe many dynamic sequences feature objects with simple, low-frequency, and often single-color textures, limiting appearance diversity. 
To address this, we randomly retexture object sequences, enriching the dataset (see~\cref{fig:augemntations}) used for the encoder/decoder training. 
Specifically, we first assign UV coordinates to the mesh of the first frame, and the UV coordinates are carried over to future frames during animation. 
This enables us to render consistent colors across frames. 
We randomly sample an image from OpenImages'~\cite{kuznetsova2020open} training split as the texture. 
Rendering via \texttt{nvdiffrast}~\cite{laine2020diffrast} operates at over 500~fps, ensuring that we can perform the retexturing on-the-fly during training.
Note that the mesh correspondences are \textit{not} provided to our model as input nor as supervision. It is merely for data augmentation and is optional.

\vspace{-1em}
\paragraph{Tracking}

To the best of our knowledge, we are the first to use dynamic Objaverse~\cite{deitke2022objaverse} as a dataset for large-scale 3D point tracking, for which we plan to release code. 
To create tracks, like our approach for texture augmentation, we use known vertex correspondences across frames to propagate the barycentric coordinates for a query point in the first frame across time. 
See the supplement for details.

\vspace{-1em}
\paragraph{Cloth simulation}
We render a dataset for cloth simulation based off~\citet{lips2024learning}, leveraging PyFlex~\cite{macklin2014unified,li2018learning}. 
First, we generate a collection of random t-shirts, shorts, and towels, each represented as a single-layer mesh. 
For each sample, we randomly select an image from OpenImage~\cite{kuznetsova2020open} as texture, randomly apply one or two folds by grabbing vertices, then lift the folded cloth to a random height and release it with zero initial velocity. 
We use unique mesh shapes and texture images for train/validation/test splits.

\begin{figure*}[htbp]
    \centering
\includegraphics[width=\textwidth]{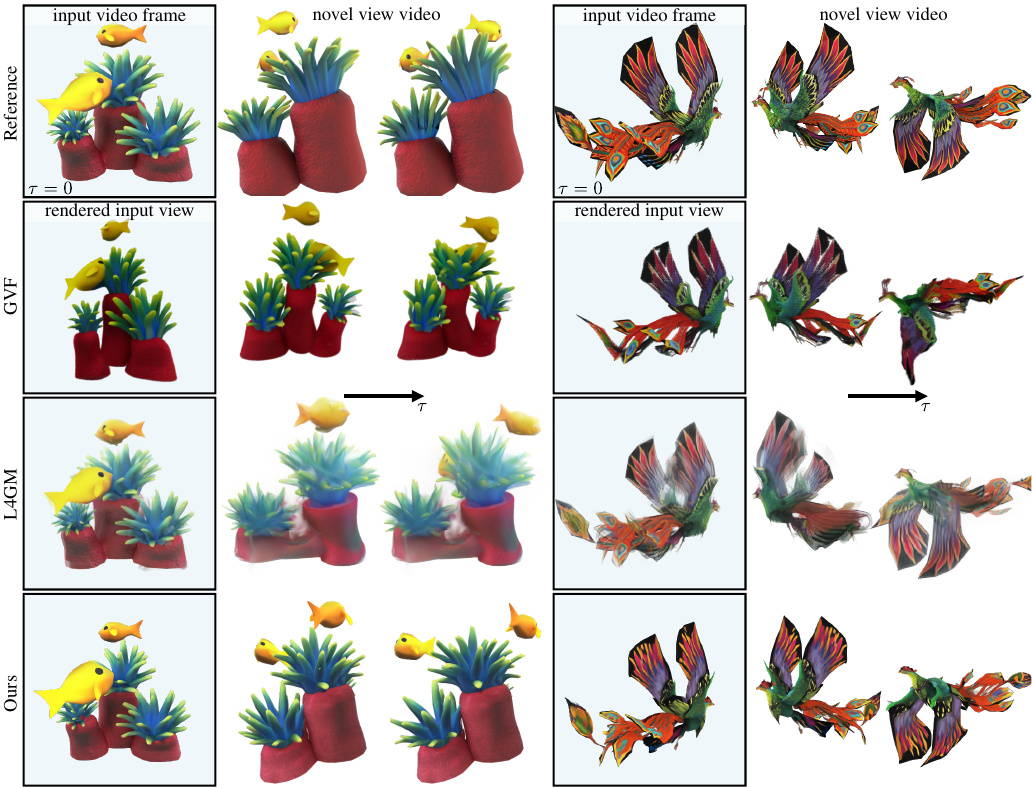}
    \vspace{-6mm}
    \caption{\textbf{Video-to-4D results.} \textbf{2nd row:} GVF~\cite{zhang2025gaussian} fails to generalize across complex shape deformations, and does not faithfully reproduce both input and novel views. \textbf{3rd row:} L4GM renders the input view faithfully due to its pixel-based Gaussian formulation and test-time alignment. However, it fails to generate realistic novel views, with visible floaters around the object. \textbf{Bottom row:} Our method recovers both accurate input views and novel view generations, even for complex structures like the bird (RHS). Asset credit~\cite{zakyora_phoenixbird,paula1221965}.}
    \label{fig:video24d_res}
\end{figure*}

\begin{figure}[htbp]
    \centering
    \includegraphics[width=\columnwidth]{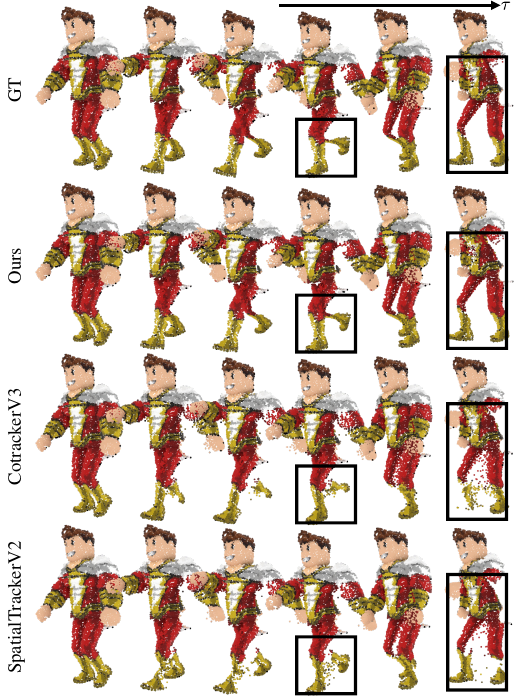}
    \vspace{-6mm}
    \caption{\textbf{3D tracking results.}  Given an input RGBD video and query points to be tracked on the first frame, we visualize the backprojected output tracks of each method. An accurate track would produce a point cloud similar to the ground-truth tracks (top row). Asset credit:~\cite{SHAZAM_ROBLOX}.}
    \label{fig:tracking_res}
    \vspace{-1em}
\end{figure}

\begin{figure}[t] 
    \centering
    \includegraphics[width=\columnwidth]{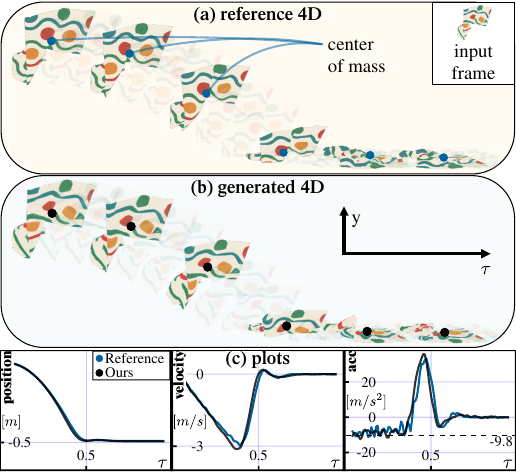}
    \vspace{-6mm}
    \caption{\textbf{Cloth simulation results.}  \textbf{(b)} Our generative model predicts the full 4D trajectory of a cloth from an initial position (\textbf{Top right:} \textit{input frame}). \textbf{(c)} Position, velocity, and acceleration (acc) of the center-of-mass point (marked in \textbf{(a)} and \textbf{(b)}) along the negative $y$ axis, plotted against time. Both the generated and reference trajectories show an initial acceleration toward the ground followed by an upward rebound after impact at $\tau \approx 0.5$.}    \label{fig:cloth_res}
    \vspace{-1em}
\end{figure}

\begin{table}[t]
\captionof{table}{\textbf{Reconstruction evaluation.} Reconstruction quality across methods, including different voxelization methods for our Gaussian decoder. Chamfer values are multiplied by $10^5$.}
\label{tab:reconstruction}
    \vspace{-0.5em}
    \centering
    \setlength{\tabcolsep}{3pt}
    \resizebox{\columnwidth}{!}{
    \begin{tabular}{lccccccc}
        \toprule
        Method & PSNR$\,\uparrow$ & SSIM$\,\uparrow$ & LPIPS$\,\downarrow$ & FVD$\,\downarrow$ & CVVDP$\,\uparrow$ & Chamfer$\,\downarrow$ \\
        \midrule
        GVF~\cite{zhang2025gaussian} & 26.45 & 0.952 & 0.051 & 173.62 & 7.620 & - \\
        LiTo~\cite{chang2026lito} (GT~vox.)& \cellcolor{tabthird}32.55 & \cellcolor{tabthird}0.972 & \cellcolor{tabthird}0.037 &  \cellcolor{tabthird}126.38 & \cellcolor{tabthird}8.544 & \cellcolor{tabthird}38.17\\
        Ours (samp.~vox.) & \cellcolor{tabsecond}34.25 & \cellcolor{tabsecond}0.981 & \cellcolor{tabsecond}0.024 &  \cellcolor{tabsecond}59.37 & \cellcolor{tabsecond}8.811 & \cellcolor{tabfirst}36.36\\
        Ours (dec.~vox.) & \cellcolor{tabfirst}35.11 & \cellcolor{tabfirst}0.983 & \cellcolor{tabfirst}0.022 &  \cellcolor{tabfirst}50.68 & \cellcolor{tabfirst}8.877 & \cellcolor{tabfirst}36.36\\

        \midrule 
                Ours (GT~vox.) & 35.39 & 0.984 & 0.021 & 48.99 & 8.910 & 36.36\\

                \midrule
        Ours-S w/o aug. & 32.41 & 0.973 & 0.036 & 99.16 & 8.469 & 43.34\\
        Ours-S w/ aug. & \cellcolor{tabfirst}33.93 & \cellcolor{tabfirst}0.979 & \cellcolor{tabfirst}0.029 & \cellcolor{tabfirst}66.78 & \cellcolor{tabfirst}8.736&\cellcolor{tabfirst}38.79\\
        \bottomrule
    \end{tabular}
    }
\end{table}

\begin{table}[t]
    \captionof{table}{\textbf{Objaverse 3D tracking evaluation}. For $L_2$ distance and $\text{APD}$, we report variants computed over all points $L_{2,\text{all}}^{3D}$/$\text{APD}_\text{all}^\text{3D}$, and only visible points $L_{2,\text{vis}}^{3D}$/$\text{APD}_\text{vis}^\text{3D}$.}
    \label{tab:tracking}
    \vspace{-0.5em}
    \centering
    \setlength{\tabcolsep}{3pt}
    \resizebox{\columnwidth}{!}{
    \begin{tabular}{lccccccccccc}
        \toprule
        Run & $L_{2,\text{all}}^{3D}\,\downarrow$ & $L_{2,\text{vis}}^{3D}\,\downarrow$ & $\text{APD}_\text{all}^\text{3D}\,\uparrow$ & $\text{APD}_\text{vis}^\text{3D}\,\uparrow$ & $\text{AJ}^\text{3D}\,\uparrow$ &   $\text{OA}\,\uparrow$\\
        \midrule
SpatialTrackerV2~\cite{xiao2025spatialtrackerv2} &  \cellcolor{tabthird}0.068 &  \cellcolor{tabthird}0.056 &  \cellcolor{tabthird}0.600 &  \cellcolor{tabthird}0.627 &  \cellcolor{tabthird}0.444 &   \cellcolor{tabfirst}0.897 \\
Cotracker3~\cite{karaev2024cotracker3} w/ GT depth       & \cellcolor{tabsecond}0.060 & \cellcolor{tabsecond}0.039 & \cellcolor{tabsecond}0.755 & \cellcolor{tabsecond}0.803 & \cellcolor{tabsecond}0.648 & \cellcolor{tabsecond}0.889 \\
Ours               &  \cellcolor{tabfirst}0.025 &  \cellcolor{tabfirst}0.020 &  \cellcolor{tabfirst}0.835 &  \cellcolor{tabfirst}0.857 &  \cellcolor{tabfirst}0.709 &  \cellcolor{tabthird}0.871\\
        \bottomrule
    \end{tabular}
    }
    \vspace{-1em}
\end{table}

\begin{table*}[t]
\captionof{table}{\textbf{Video-to-4D generation evaluation.} For quantitative metrics, we render the generated 4D objects from 10 viewpoints. We report separate metrics for the input and 9 novel viewpoints on Consistent4D~\cite{jiang2024consistentd} (C4D) and 128 test objects in Objaverse~\cite{deitke2022objaverse} (Obj). 
The input video is rendered from random distance to the origin ($r \in [2, 5]$) with a paired field of view to cover the entire object.
}

\centering
\setlength{\tabcolsep}{6pt}
\footnotesize


\begin{tabular}{llcccccccc}
\toprule
 & &
\multicolumn{4}{c}{Conditioning View} & 
\multicolumn{4}{c}{Novel View} \\
\cmidrule(lr){3-6} \cmidrule(lr){7-10}
& Method & PSNR$\uparrow$ & LPIPS$\downarrow$ & FVD$\downarrow$ & CLIP$\uparrow$
& PSNR$\uparrow$ & LPIPS$\downarrow$ & FVD$\downarrow$ & CLIP$\uparrow$
\\
\midrule
\multirow{3}{*}{\rotatebox{90}{\shortstack{Obj}}}
& L4GM~\cite{ren2024l4gm} & \cellcolor{tabsecond}22.59 ± 4.74 & \cellcolor{tabsecond}0.074 ± 0.058 & \cellcolor{tabfirst}184 ± 180 & \cellcolor{tabsecond}0.95 ± 0.04 & \cellcolor{tabsecond}18.30 ± 4.85 & \cellcolor{tabsecond}0.146 ± 0.072 & \cellcolor{tabsecond}498 ± 261 & \cellcolor{tabsecond}0.90 ± 0.04 \\
& GVF~\cite{zhang2025gaussian} & 18.21 ± 4.90 & 0.117 ± 0.053 & 571 ± 390 & 0.89 ± 0.05 & 16.41 ± 4.73 & 0.157 ± 0.064 & 663 ± 327 & 0.88 ± 0.04 \\
& Ours & \cellcolor{tabfirst}24.04 ± 5.76 & \cellcolor{tabfirst}0.056 ± 0.032 & \cellcolor{tabsecond}245 ± 378 & \cellcolor{tabfirst}0.96 ± 0.03 & \cellcolor{tabfirst}20.62 ± 6.04 & \cellcolor{tabfirst}0.104 ± 0.052 & \cellcolor{tabfirst}373 ± 307 & \cellcolor{tabfirst}0.94 ± 0.03 \\
\midrule

\multirow{3}{*}{\rotatebox{90}{\shortstack{C4D }}}
& L4GM~\cite{ren2024l4gm} & \cellcolor{tabsecond}21.87 ± 5.46 & \cellcolor{tabsecond}0.080 ± 0.053 & \cellcolor{tabfirst}118 ± 46 & \cellcolor{tabfirst}0.98 ± 0.01 & \cellcolor{tabsecond}17.71 ± 2.39 & \cellcolor{tabsecond}0.147 ± 0.060 & \cellcolor{tabsecond}432 ± 202 & \cellcolor{tabsecond}0.93 ± 0.04 \\
& GVF~\cite{zhang2025gaussian} & 17.76 ± 4.91 & 0.122 ± 0.076 & 390 ± 160 & 0.93 ± 0.04 & 15.80 ± 3.49 & 0.163 ± 0.078 & 584 ± 216 & 0.92 ± 0.03 \\
& Ours & \cellcolor{tabfirst}22.95 ± 4.43 & \cellcolor{tabfirst}0.072 ± 0.046 & \cellcolor{tabsecond}188 ± 143 & \cellcolor{tabsecond}0.98 ± 0.01 & \cellcolor{tabfirst}18.98 ± 3.18 & \cellcolor{tabfirst}0.120 ± 0.057 & \cellcolor{tabfirst}431 ± 198 & \cellcolor{tabfirst}0.94 ± 0.02 \\
\bottomrule
\end{tabular}
\label{tab:generation}
\vspace{-1em}
\end{table*}

\section{Results}

\subsection{Reconstruction}

We evaluate the reconstruction quality of our representation, comparing against LiTo~\cite{chang2026lito}, a state-of-the-art 3D representation for which we feed in the point cloud frame-wise, and GVF~\cite{zhang2025gaussian}, a method which learns a latent space for 4D deformations.  
We finetune LiTo to handle a smaller set of latents per frame such that the total latent size for the entire sequence matches our technique. 
For \cref{tab:reconstruction}, we compute metrics on a held out set of 256 scenes from Objaverse. 
Please see \cref{fig:recon_res} for qualitative results.

\vspace{-1em}
\paragraph{Metrics.} To evaluate appearance, we compute image metrics on the rendered output 3D Gaussians of each method, such as PNSR, SSIM and LPIPS~\cite{zhang2018perceptual} (calculated using VGG~\cite{simonyan2015deepconvolutionalnetworkslargescale} features). We also employ the video metrics FVD~\cite{unterthiner2018towards} (using Inception-v1 I3D~\cite{carreira2017quo}) and CVVDP~\cite{mantiuk2024color} to evaluate temporal consistency. For geometry, we use the symmetric Chamfer distance between the ground-truth and sampled point cloud, calculated on 8000 points.

\vspace{-1em}
\paragraph{The importance of joint temporal encoding.} 
To investigate the significance of spatiotemporal processing, we evaluate LiTo~\cite{chang2026lito}, which processes each frame independently. Quantitatively, its reconstruction quality across geometry and appearance is lower than ours. 
Visual comparisons reveal prominent temporal inconsistencies that show
jointly processing the point cloud across time is crucial.

\vspace{-1em}
\paragraph{How do the voxels provided to the Gaussian decoder affect the reconstruction?}
We evaluate our method (trained with GT voxels) on reconstruction, given three different methods to extract the voxels, a) GT voxels, b) voxels derived from $65536$ points sampled from the 4D surface decoder per frame, and c) voxels from the trained voxel decoder. 
\cref{tab:reconstruction} shows there is a drop off in quality when using the sampled points (labeled ``samp.~vox.'') ---we believe this is because insufficient points are sampled to create accurate sparse voxels. On the other hand, our trained voxel decoder (``dec.~vox.'') gives similar performance to GT voxels, while being far faster than the sampling approach (timing in supplement).

\vspace{-1em}
\paragraph{Modeling deformations vs. time-conditioned 3D objects.} 
We also compare against the current SoTA 4D reconstruction method, GVF~\cite{zhang2025gaussian}, which models deformations rather than time-conditioned geometry and appearance. Our approach outperforms GVF in both reconstructed appearance and geometry. We believe this is because modeling deformations only may be insufficient for the complexities of real dynamic scenes, \eg, scenes where components appear or disappear cannot be easily handled.

\vspace{-1em}
\paragraph{Effect of texture augmentation.} 
To assess the impact of our texture augmentation, we train smaller versions of our model with (Ours-S w/ aug.) and without texture augmentation (Ours-S w/o aug.). As shown in \cref{tab:reconstruction}, the texture diversity in the unaugmented dynamic dataset is insufficient to capture high-frequency details, resulting in significantly lower appearance reconstruction metrics. Interestingly, we also observe a large degradation in geometry performance.

\subsection{Video-to-4D}
\paragraph{Metrics and baselines.} To evaluate our video-to-4D model, we report metrics separately for the input view and for 9 novel views rendered from 4D representations. For both, we compute FVD (using Inception-v1 I3D~\cite{carreira2017quo}), LPIPS, PSNR, and CLIP similarity~\cite{radford2021clip}. We compare against GVF~\cite{zhang2025gaussian} and L4GM~\cite{ren2024l4gm}, which directly predicts pixel-aligned Gaussians without a 4D generative model. Evaluation is performed on both the Consistent4D dataset~\cite{jiang2024consistentd} and an unseen dynamic Objaverse test set.

We evaluate three versions of each dataset: one with the camera at a distance of $4$ from the origin (matching GVF’s training), one at $3$ (matching L4GM’s training), and one at a random distance. 
We report values for random distances in \cref{tab:generation} (see supplement for full results). 

\vspace{-1em}
\paragraph{Comparison.} GVF performs well on the Consistent4D dataset but struggles to generalize beyond it, likely due to the smaller training set and the challenges of modeling complex deformations in Objaverse. 
On the other hand, benefiting from its pixel-aligned Gaussians, L4GM achieves high accuracy for the input view, especially when $r=3$ (its training setting), but struggles to produce sharp novel views due to the absence of direct 4D modeling. 
In comparison, our method outperforms both baselines in input view reconstruction and novel view generation, and generalizes more robustly across different camera settings, which we attribute to the strong representation learned by our encoder.

\subsection{3D tracking}
\paragraph{Metrics and baselines.} 
We evaluate our 3D tracking with the $\text{APD}^{3D}$, $\text{AJ}^{3D}$, and $\text{OA}$ metrics from Koppula et al.~\cite{koppula2024tapvid3d}, as well as 3D $L_2$ distance. For $L_2$ and $\text{APD}^{3D}$, we report variants computed over all points $L_{2,\text{all}}^{3D}$/$\text{APD}_\text{all}^\text{3D}$, and only visible points $L_{2,\text{vis}}^{3D}$/$\text{APD}_\text{vis}^\text{3D}$. 

We compare against SpatialTrackerV2~\cite{xiao2025spatialtrackerv2}, a recent method for 3D tracking which uses dual 2D and 3D tracking heads; and CoTracker3~\cite{karaev2024cotracker3}, for which we use ground-truth depth to reproject the tracks into 3D. 
See \cref{fig:tracking_res} and \cref{tab:tracking} for the full set of results. During metric calculation, we remove all points that lie outside scene bounds for CoTracker3 and SpatialTrackerV2.

\vspace{-1.em}
\paragraph{Comparison.} Surprisingly, CoTracker3, a 2D tracking method, outperforms SpatialTrackerV2, likely because object tracking in our setup involves fewer occlusions and the 2D tracks produced by CoTracker3 are highly accurate. 
SpatialTrackerV2, on the other hand, exhibits numerous flying-pixel artifacts, particularly at depth discontinuities (see \cref{fig:tracking_res}). 
Our method surpasses both approaches, with a notable strength in tracking textureless objects and regions. 
%
%
This advantage arises from the fact that our representation models dynamic geometry and appearance jointly. 
However, our occlusion accuracy is lower than the other methods because we do not directly predict occlusions. 

\subsection{Cloth simulation}

\cref{fig:cloth_res} presents an illustrative example of our generated sequence. 
The physical behavior of the generated object---its position, velocity, and acceleration over time---closely follows that of the reference. 
The plots highlight interesting characteristics, such as the sharp upward acceleration of the center of mass after contacting the ground (indicating rebound), and the corresponding abrupt reversal in velocity. 
Quantitatively, over our test set, our predicted trajectories achieve a RMSE in center-of-mass position of $1\text{cm}$ (0.5\% of the maximum scene size), showing its precision. 
Please see the supplement for a set of full metrics on rendered cloth videos and center-of-mass trajectories.

We emphasize the difficulty and under-constrained nature of this task: the generative model must (a) infer the object’s shape, (b) predict its deformation while moving through the air, and (c) model its deformation upon collision. 
The fact that our model produces physically realistic quantities highlights the strength of our representation, which enables such structured generation under limited constraints.


\section{Discussion}

Our work introduces a method to learn a compact 4D representation that jointly models time-varying geometry and appearance without relying on temporal correspondences as encoder input. 
We enable these advancements through innovations including data augmentation, careful architecture designs like 4D spatiotemporal patchification to improve memory efficiency, and joint supervision of 4D geometry and appearance. 
Our evaluations show that \dst reconstruct dynamic scenes more faithfully and efficiently than deformation-based or per–time-step alternatives, and that the learned tokens transfer effectively to video-to-4D generation, 3D tracking, and cloth simulation. 

\vspace{-1em}
\paragraph{Limitations.} 
Our joint spatiotemporal encoding reduces temporal instability, but residual flickering can appear in regions with complex textures or large motions. 
This primarily stems from the decoder operating on each output frame independently, a constraint imposed by GPU memory. 
Designing a more memory-efficient decoder capable of processing multiple timestamps of Gaussians jointly could alleviate these artifacts.
Texture quality in video-to-4D generation can be lower when the input video contains high-frequency detail. 
This appears to arise from the limited capacity of the underlying generative model---which is of similar size as existing 3D generative approaches~\cite{chang2026lito,xiang2024structured}---as indicated by the gap between our reconstruction and generation performance.
Finally, training \dst currently requires substantial computational resources and long training times. 
The absence of large-scale 4D foundation models to initialize from contributes significantly to this cost; future pretraining efforts could greatly reduce the overhead. 
We hope that releasing our model and codebase will support future work toward more efficient training and broader 4D representation learning and applications.

\section*{Acknowledgements}
Work was done during AM's internship at Apple. We are grateful to Hadi Pouransari, Stephen Pulman, and Wei-Yu Chen for providing helpful comments on the manuscript.

{
    \small
    \bibliographystyle{ieeenat_fullname}
    \bibliography{main}
}

\clearpage
\setcounter{page}{1}
\maketitlesupplementary

\begin{table*}[ht]
  \centering
      \captionof{table}{\textbf{Overview of related 4D representations.}}

\label{tab:related_work}
  \resizebox{\textwidth}{!}{
  \begin{tabular}{lcccccc}
    \toprule
    \textbf{name} & \textbf{geometry} & \textbf{appearance} & \textbf{pretrained model use} & \textbf{total latent dimension} & \textbf{input to encoder} & \textbf{training dataset} \\
    \midrule
    ShapeGen4D~\cite{yenphraphai2025shapegen} & SDF & - & \checkmark & not reported & point cloud, tracks & Objaverse, ObjaverseXL \\

    Gaussian Variational Fields~\cite{zhang2025gaussian} & only 3DGS & 3DGS & \checkmark & $20,000 \times 11 + \tau \times 512 \times 16$ & point cloud, tracks, SLAT~\cite{xiang2024structured} & Objaverse, ObjaverseXL \\
    \rowcolor{blue!5}
    \textbf{Ours} & $p(x,y,z \mid t)$ & 3DGS & - & $8192 \times 32$ & point cloud & Objaverse, ObjaverseXL \\
    \midrule
    Cut3r~\cite{wang2025cut3r} & pointmap & image & \checkmark & $\tau \times 768 \times 768$  & video & 32 datasets \\
    4DS~\cite{carreira2024scaling4drepresentations} & - & - & - & - & video & proprietary dataset \\
    \midrule
    CasPeR~\cite{rempe2020caspr} & $p(x,y,z \mid t)$ & - & - & 1600 & point cloud & ShapeNet \\

    Motion2VecSets~\cite{wei2024motion} & occupancy field & - & - & $\tau \times 512 \times 512$ & point cloud & DT4D~\cite{wei2024motion} \\
    Hyperdiffusion~\cite{erkoç2023hyperdiffusion} & SDF & - & - & $32,000$ & point cloud, initialization weights & DT4D~\cite{wei2024motion} \\
    DNF~\cite{zhang2024dnf} & SDF & - & - & $384$ & point cloud & DT4D~\cite{wei2024motion} \\

    \bottomrule
  \end{tabular}
  }
\end{table*}

\begin{table*}[h]
    \captionof{table}{\textbf{Tracking evaluation.} Comparison of 2D and 3D tracking accuracy across methods. We show results with filtering (filt.), which involves removing points outside the bounding box $[-1, 1]^3$. }
    \label{tab:supp_tracking}
    \vspace{-0.5em}
    \centering
    \setlength{\tabcolsep}{4pt}
    \resizebox{\textwidth}{!}{
    \begin{tabular}{lcccccccccccc}
        \toprule
        Run & Filt. & $L_2^{3D}\,\downarrow$ & $L_{2,\text{vis}}^{3D}\,\downarrow$ & $\text{APD}^\text{3D}\,\uparrow$ & $\text{APD}_\text{vis}^\text{3D}\,\uparrow$ & $\text{AJ}^\text{3D}\,\uparrow$ & $L_2^{2D}\,\downarrow$ & $L_{2,\text{vis}}^{2D}\,\downarrow$ & $\text{APD}^\text{2D}\,\uparrow$ & $\text{APD}_\text{vis}^\text{2D}\,\uparrow$ &  $\text{AJ}^\text{2D}\,\uparrow$ & $\text{OA}\,\uparrow$\\
        \midrule
Cotracker3        & - &  \cellcolor{tabthird}0.143 &  \cellcolor{tabthird}0.109 & \cellcolor{tabsecond}0.736 & \cellcolor{tabsecond}0.784 & \cellcolor{tabsecond}0.649 & \cellcolor{tabsecond}5.651 & \cellcolor{tabsecond}4.457 & \cellcolor{tabsecond}0.730 & \cellcolor{tabsecond}0.760 & \cellcolor{tabsecond}0.616 & \cellcolor{tabsecond}0.884 \\
Spatial Tracker V2 & -& \cellcolor{tabsecond}0.121 & \cellcolor{tabsecond}0.107 &  \cellcolor{tabthird}0.589 &  \cellcolor{tabthird}0.616 &  \cellcolor{tabthird}0.445 &  \cellcolor{tabthird}5.911 &  \cellcolor{tabthird}4.752 &  \cellcolor{tabthird}0.723 &  \cellcolor{tabthird}0.752 &  \cellcolor{tabthird}0.597 &  \cellcolor{tabfirst}0.897 \\
Ours               & - &  \cellcolor{tabfirst}0.025 &  \cellcolor{tabfirst}0.020 &  \cellcolor{tabfirst}0.835 &  \cellcolor{tabfirst}0.857 &  \cellcolor{tabfirst}0.709 &  \cellcolor{tabfirst}3.788 &  \cellcolor{tabfirst}3.100 &  \cellcolor{tabfirst}0.772 &  \cellcolor{tabfirst}0.793 &  \cellcolor{tabfirst}0.631 &  \cellcolor{tabthird}0.871\\

\midrule
Cotracker3         & \checkmark & \cellcolor{tabsecond}0.060 & \cellcolor{tabsecond}0.039 & \cellcolor{tabsecond}0.755 & \cellcolor{tabsecond}0.803 & \cellcolor{tabsecond}0.648 & \cellcolor{tabsecond}5.295 & \cellcolor{tabsecond}4.169 & \cellcolor{tabsecond}0.736 & \cellcolor{tabsecond}0.765 & \cellcolor{tabsecond}0.606 & \cellcolor{tabsecond}0.889 \\
Spatial Tracker V2 & \checkmark &  \cellcolor{tabthird}0.068 &  \cellcolor{tabthird}0.056 &  \cellcolor{tabthird}0.600 &  \cellcolor{tabthird}0.627 &  \cellcolor{tabthird}0.444 &  \cellcolor{tabthird}5.874 &  \cellcolor{tabthird}4.722 &  \cellcolor{tabthird}0.723 &  \cellcolor{tabthird}0.752 &  \cellcolor{tabthird}0.585 &  \cellcolor{tabfirst}0.897 \\
        \bottomrule
    \end{tabular}
    }
\end{table*}

\paragraph{Overview of the supplemental material.}

\begin{itemize}[leftmargin=*]
\item \cref{sec: supp dataset} discusses details about our data preparation.
\item \cref{sec: supp model info} provides details about model architectures and training, \eg, learning rates, total iterations, and runtime.
\item \cref{sec: supp additional results} includes more results on video-to-4D generation, 3D tracking, cloth simulation and real-world experiments.
\item \cref{sec: supp additional related work} introduces additional related works in the area of 3D tracking and 4D generation.
\item \cref{sec:author_contrib} explains the individual author contributions.

\end{itemize}

\section{Additional dataset details}
\label{sec: supp dataset}
\paragraph{Objaverse preprocessing.} 
As a preprocessing step, we normalize the mesh animation sequence into a bounding box of $[-1, 1]^3$, using the dependency graph in Blender.
We render videos of each object from $35$ viewpoints, with the cameras placed uniformly on a sphere of radius $3.5$ centered at the origin.
The cameras look at the world origin and are with a field of view of 40 degrees.
We also render videos of the scene from a separate set of $25$ viewpoints, where the camera positions lie randomly from $1.5$ to $3.5$ units away from the center, looking at the random position within a centered ball of radius $0.25$, and with field of view randomly selected from $25$ to $40$ degrees.
We rendered RGBD images of resolution $532\times532$.
The first set of images is used to create the input point clouds, so we turn off anti-aliasing during rendering; the other is used as supervision for the Gaussian decoder, and we set the anti-aliasing filter width to 1 pixel.

To train the video-to-4D model, we additionally render the conditioning videos by placing cameras at a random distance from $1.75$ to $5$ units to the origin.
The cameras look at the origin and with field of view set to cover $[-1, 1]^3$:
\begin{equation}
\text{fov} = 2 \times \arctan\left(\frac{1.25}{r - 0.5}\right) \text{radians}.
\end{equation}
These videos are anti-aliased with filter width of 1 pixel, and during training, we rotate the world coordinate (along with the corresponding input point cloud) so the conditioning cameras always lie on the +x axis. This operation enables our generations to align with the input videos.

\paragraph{3D tracking.}
We train our tracking model to only take as input the points that are visible in the first RGBD video frame.
There are two options: 1) backproject the first RGBD frame and identify the intersected triangles, gather barycentric coordinates, and use them to compute tracks, and 2) randomly sample on the mesh surfaces uniformly and keep only the closest points to the back-projected RGBD pixels from the first frame. 
We tried both approaches and found no significant differences, so we choose the second method as it is faster to compute on-the-fly.

\begin{table*}[h]

\newcommand{\threemulti}[1]{\multirow{3}{*}{#1}}
\caption{\textbf{Model size and training details.} The table overviews architecture and training details about individual models. $\searrow$ indicates the learning rate decays according to the schedule used by \citet{vaswani2017attention}, with the peak learning rate shown in the table and 4000 warm-up iterations.}
    \label{tab:model_info}
\centering
\footnotesize
\begin{tabular}{l l l c c r r r }
\toprule
\textbf{} & 
\textbf{model} & 
\textbf{dataset / note} & 
\textbf{parameters} ($\times 10^6$) &
\textbf{lr ($\times 10^{-4}$)} &
\textbf{batch} &
\textbf{iters} ($\times 10^3$) &
\textbf{H100 hours ($\times 10^3$)}  \\
\midrule
1-1 & encoder            &  Objaverse & 58.6 & \threemulti{4.9 $\searrow$} & \threemulti{256} & \threemulti{132}   & \threemulti{15.4}  \\
1-2 & 4D surface decoder & Objaverse & 12.2 &        &    &      &        \\
1-3 & Gaussian decoder   & Objaverse & 75.8 &        &    &      &      \\
\midrule
2-1 & encoder (finetune from 1-1) &  synth-cloth  & 58.6 & \threemulti{4.9 $\searrow$} & \threemulti{256} & \threemulti{80}    & \threemulti{3.8}  \\
2-2 & 4D surface decoder  & synth-cloth & 12.2 &        &    &      &        \\
2-3 & Gaussian decoder    &  synth-cloth & 75.8 &        &    &      &       \\
\midrule
3 & voxel decoder (for 1-1)      & Objaverse  & 65.0 & 4.9 $\searrow$ & 256 & 98    & 17.5 \\
4 & voxel decoder (for 2-1)      & synth-cloth  & 65.0 & 4.9 $\searrow$ & 256 & 20    & 3.8 \\
5   & 3D tracking       & Objaverse   &  123    & 1.0   & 128 & 372   & 13.2 \\
6   & video-to-4D       &  Objaverse  & 623  & 1.0   & 256 & 362.5 & 61.4  \\
7   & image-to-4D       &  synth-cloth  & 623  & 1.0   & 256 & 250   & 38.1 \\
\bottomrule
\end{tabular}
\end{table*}

\section{Model information}
\label{sec: supp model info}

Please see \cref{tab:model_info} for an overview of the model and training details. 
Below, we describe the architectures in detail.

\subsubsection{Encoder} 
Inspired by recent work in 3D representations~\cite{chang2026lito,zhang20233dshape2vecset}, we aim to encode patches of the 4D spatiotemporal surface into \dst. 
Following the encoder design of LiTo~\cite{chang2026lito}, we use a Perceiver IO~\cite{perceiverio} architecture as our encoder, where we use $k$ points sampled from the input point cloud as queries $\mathbf{x}$.  
Each of the queries attends only to a local neighborhood of the input points that are closest to the query in 4D. 
The network outputs the dynamic shape token $\textbf{s}~=~\mathcal{E}(\mathcal{X},~\textbf{q}),$ where $\mathbf{s}$ contains $k$ tokens of dimension $d$. We describe these aspects in more detail below.

\paragraph{Input point cloud.} 

The input point cloud contains points randomly sampled from the backprojected multiview RGBD videos of 24 frames. 
We randomly select from all frames, so some time instances can have more or fewer points.
The 3D coordinates are normalized to lie within a unit cube (\([-1, 1]^3\)), while temporal values assume a fixed frame rate of 100\,fps, such that each point’s time coordinate $\tau$  corresponds to \(\text{frame\_num}/100\). We sample a total of 1.44 million points, which is significantly fewer points per frame than in the 3D representation LiTo ($\sim$60 thousand \vs $\sim$1 million)~\cite{chang2026lito}. 
This makes it essential for the network to effectively share and propagate information across frames.

\vspace{-1em}
\paragraph{Query tokens.} 
Prior work in 3D representation learning that used Perceiver IO or similar architectures have typically obtained query inputs through learned tokens~\cite{zhang20233dshape2vecset, chang20243dshapetokenization} or sampling from input points~\cite{qi2017pointnetplusplus, zhang20233dshape2vecset}. 
We adopt furthest point sampling~\cite{qi2017pointnetplusplus} in the 4D space, such that the queries correspond to spatio-temporal points.
This design allows each query to encode richer scene information by incorporating both the geometric and temporal contexts.
Following \cite{li2025triposg}, we first  select $4 \times k$ points randomly and apply farthest point sampling on the selected $4\times k$ points to get $k$ points.

\vspace{-1em}
\paragraph{Attention design.} 
To improve computational efficiency, we use voxel-based windowed self-attention, similar to prior work~\cite{chang2026lito}. 
Specifically, only tokens within the same 4D voxel attend to each other. 
Our voxelization scheme accounts for both spatial and temporal dimensions --- along $x, y, z$ the voxel cell width is $0.5$ (out of 2) and along $t$ the voxel cell width is 6 frames (out of 24 frames).

For cross-attention, each query point only attends to the input points that are closest to itself. 
This is the extension of the 3D patchification~\cite{chang2026lito}
 in 4D.
Both our cross- and self-attention can be implemented efficiently with xformers~\cite{xFormers2022} or flash-attention~\cite{flashattention}.

\begin{figure}[htbp]
    \centering
    \includegraphics[width=\columnwidth]{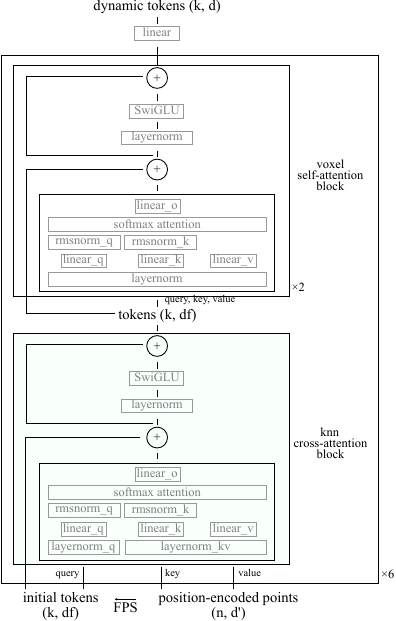}
    \vspace{-6mm}
    \caption{\textbf{Encoder archtecture.}}
    \label{fig:arch_tokenizer}
    \vspace{-1em}
\end{figure}

\paragraph{Architecture.} 
See \cref{fig:arch_tokenizer} for the architecture of the encoder. 
Our encoder employs a Perceiver-style point encoder tailored for spatiotemporal point clouds. 
We initialize a latent set of 8,192 queries using FPS-based subsampling and process them with two Perceiver blocks, each comprising a localized cross-attention layer followed by six localized self-attention layers. 
Both attention types operate with 16 heads and a 512-dimensional latent space. 
Input tokens are a concatenation of Fourier positional encoding of 3D coordinates, RGB values, and timestamps.
All feed-forward layers use SwiGLU~\cite{shazeer2020swiglu} activations with RMSNorm~\cite{zhang2019rmsnorm}. 
After the attention stack, the latent set is projected to a 32-dimensional representation used as the \dst.

\subsubsection{4D surface decoder} 
\begin{figure}[htbp]
    \centering
    \includegraphics[width=\columnwidth]{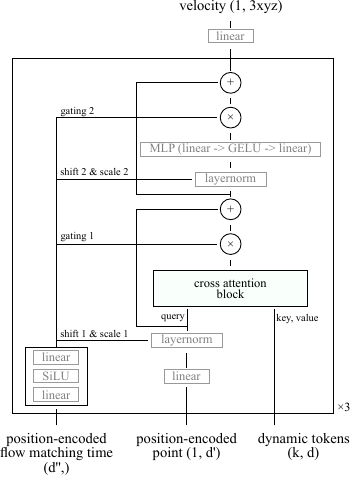}
    \vspace{-6mm}
    \caption{\textbf{4D surface decoder archtecture.}}
    \label{fig:arch_vel}
    \vspace{-1em}
\end{figure}

We use the decoder architecture from \citet{chang20243dshapetokenization} and further condition the model on the Fourier encoded frame time $\tau$. 
To ensure that we model a 4D distribution \(p(\mathbf{x} \vert \tau, \mathbf{s})\), the decoder operates independently on each point, using only cross-attention and point-wise operations, \ie, no self-attention between points.
See~\cref{fig:arch_vel} for the architecture of the 4D surface decoder.

\subsubsection{Gaussian decoder} 
Following recent 3D representations~\cite{chang2026lito}, we use a Perceiver IO~\cite{perceiverio} architecture for our Gaussian decoder \(G\). 
Coarse occupied voxel centers's 4D coordinates `$\text{Vox}$' are used as the queries, 
which attend to the \dst $\mathbf{s}$. 
For each occupied voxel, the decoder predicts 64 Gaussians, each is 14-dimensional, including position, scale, rotation modeled as a quaternion, opacity, and diffused RGB.
During training, we use ground-truth sparse occupancy voxels. 
During test time, one option is to directly sample point clouds using the 4D surface decoder to construct the voxel grid, but this involves integrating an ODE on a large number of points, and thus is computationally costly. 
To tackle this, we train a separate voxel decoder that predicts occupancy voxels directly from the pretrained \dst, following a similar architecture to the sparse structure network used by Trellis~\cite{xiang2024structured}. Specifically, the voxel decoder predicts Trellis'~\cite{xiang2024structured} occupancy latent (OL), which we supervise with the Huber loss. The OL is decoded by Trellis'{} pretrained decoder to get a $64^3$ occupancy probability, which we threshold to obtain a sparse voxel grid. The training is fast (${>}99\%$ accuracy within 10k iterations) and generalizes (see \cref{fig:real_world}).
This enables much faster 3D Gaussian decoding with a minor loss in quality, as shown in the paper.

\paragraph{Architecture.} 
Our Gaussian decoder maps latent shape tokens to a set of 3D Gaussians that parameterize geometry and appearance. 
The decoder conditions on voxel features comprising coordinates, time, and their Fourier embeddings, processed through a lightweight one-layer MLP. It then applies a Perceiver-style module with a 512-dimensional latent space, six blocks, global cross-attention, and two voxel-based windowed self-attention layers (eight heads each) with SwiGLU MLP. 
At the output, individual tokens corresponding to each occupied voxel are expanded by an MLP to predict the parameters for 64 Gaussians, including position offsets, pre-normalized quaternions, anisotropic scales, opacity logits, and diffused RGB coefficients. 
The decoder applies sigmoid activations for opacity and scale. 
The Gaussian scale after sigmoid is shifted and rescaled to $0.001$ and $0.01$.

\subsubsection{Cloth simulation (image-to-4D)}

\paragraph{Architecture.} We use a DiT-style diffusion transformer to model the conditional denoising process. The network operates on patch size one and receives a 2,065-dimensional conditioning token that fuses pretrained visual features, learned linear projections, and a time embedding. The model adopts the `large' configuration with 28 transformer blocks, each using a 1,152-dimensional hidden size, 16 self-attention heads, RMSNorm, and an MLP with expansion ratio 4. Residual pathways use zero drop-path. The model employs learned positional embeddings and a Fourier time embedding of 64 frequencies. 

\subsection{Runtime analysis}

Please see the runtime analysis in~\cref{tab:timings}. 
As can be seen, both reconstruction and tracking run efficiently --- encoding a dynamic point cloud containing 1.44 million points to \dst of 8192 tokens takes about $\sim$400 ms.
While decoding 3D Gaussians of one frame from \dst takes $\sim$33 ms. 

Sampling \dst from the video-to-4D model with $20$ Euler steps takes about 13 seconds. 
As we simply encode individual frames separately with DINOv2 (\ie, no temporal downsampling), most of the runtime is spent on the cross attention between the 8192 latent tokens and the 34656 ($38 \times 38 \time 24$) DINO image patch tokens.
In comparison, L4GM~\cite{ren2024l4gm} and GVF~\cite{zhang2025gaussian} take $\sim$2.4 and $\sim$35 seconds respectively.
Utilizing a video encoding instead of per-frame encoded DINO feature and utilizing recent techniques like MeanFlow~\cite{geng2025mean}, which can significantly reduce the number of numerical integration step, can reduce our runtime, and are left as future work.

For 3D tracking, our method takes $\sim$400 ms (including encoding the RGBD point cloud into shape tokens). On the other hand, 
CoTracker3~\cite{karaev2024cotracker3} and SpatialTrackerV2~\cite{xiao2025spatialtrackerv2} take 1.86 and 5.15 seconds, respectively.

\begin{table}[th]
\caption{\textbf{Runtime analysis of various use cases.}}
\label{tab:timings}
\vspace{-3mm}
\centering
\begin{tabular}{lc}
\toprule
\multicolumn{2}{l}{\textbf{reconstruction from dynamic rgb point cloud}} \\
\quad run encoder & 387.66 ms  \\
\quad voxel decoder (1 frames)  & 20.14 ms \\
\quad gaussian decoder (1 frame) & 32.71 ms \\
\quad 4D surface decoder (16384 points) & 24.80 s  \\

\midrule
\multicolumn{2}{l}{\textbf{video to 4D}} \\
\quad flow matching (20 Euler steps) & 12.87 s \\
\quad voxel decoder (1 frames) &  20.14 ms\\
\quad gaussian decoder (1 frame) & 32.71 ms \\
\midrule
\multicolumn{2}{l}{\textbf{3D tracking}} \\
\quad run encoder & 389.46 ms \\
\quad 3D tracking network & 19.65 ms \\

\bottomrule

\end{tabular}
\end{table}

\section{Additional results}
\label{sec: supp additional results}

\begin{figure*}[htbp]
    \centering
    \includegraphics[width=\textwidth]{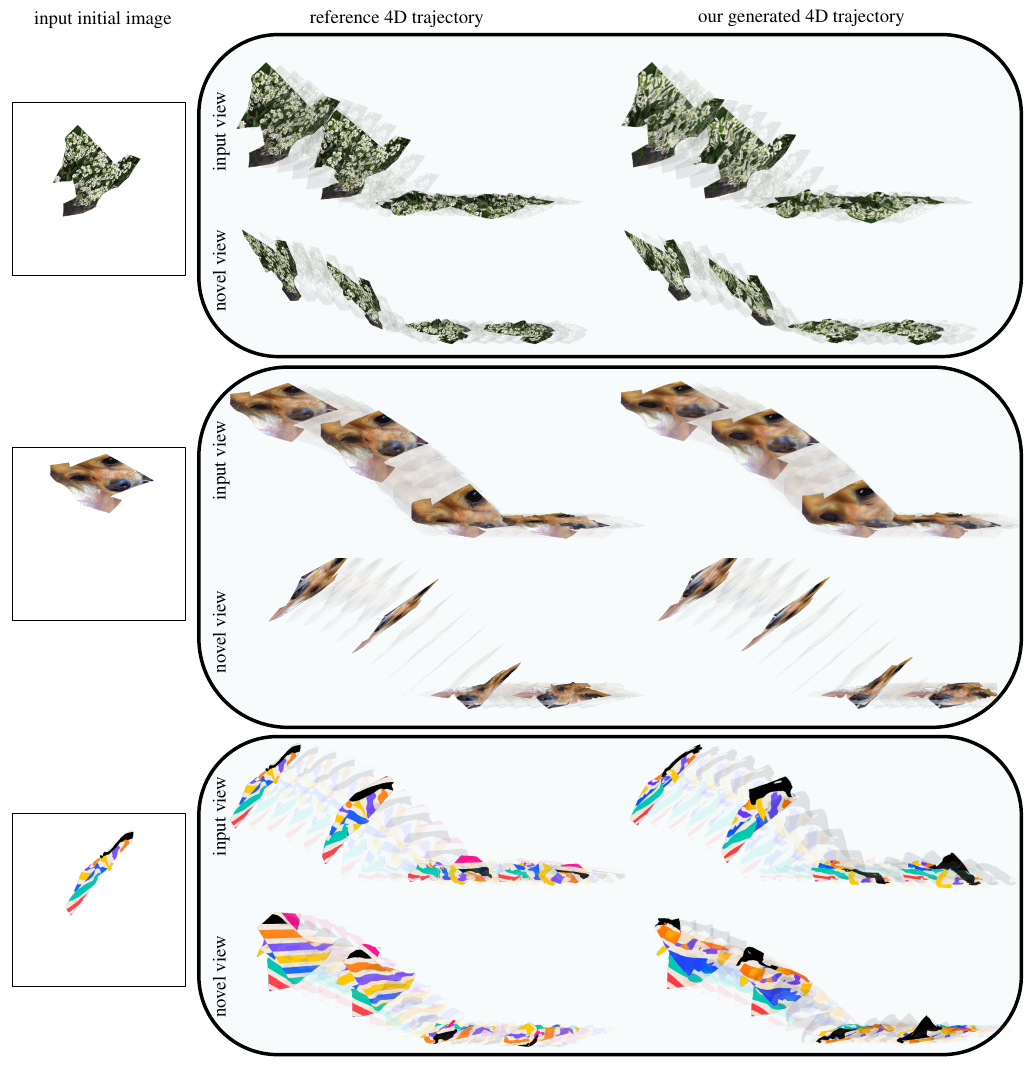}
    \vspace{-6mm}
    \caption{\textbf{Cloth simulation.} Given an input cloth initial image, we simulate the 4D cloth trajectory.}
    \label{fig:supp_cloth}
    \vspace{-1em}
\end{figure*}

\newcommand{\paneltag}[1]{%
  \rlap{\raisebox{-3.0ex}[0pt][0pt]{\hspace{0.2em}\scriptsize\bfseries(#1)}}%
}

\begin{figure*}[htbp]
    \centering
    \includegraphics[width=\textwidth]{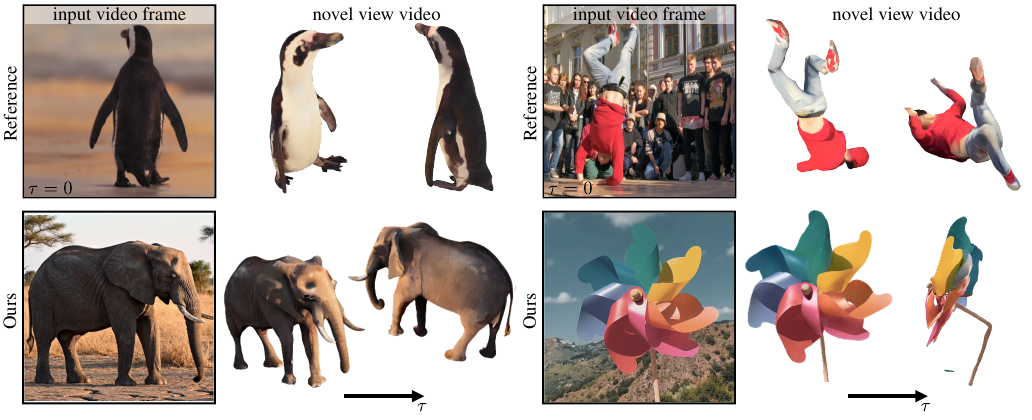}
    \vspace{-6mm}
    \caption{\textbf{Real-world generation.} We mask out real-world videos for the object of interest, and generate video-conditioned 4D objects.}
    \label{fig:real_world}
    \vspace{-1em}
\end{figure*}

\begin{table*}[t]
\captionof{table}{\textbf{Video-to-4D generation evaluation.} For quantitative metrics, we render the generated 4D objects from 10 viewpoints. We report separate metrics for the input and 9 novel viewpoints on Consistent4D~\cite{jiang2024consistentd} (C4D) and 128 test objects in Objaverse~\cite{deitke2022objaverse} (Obj). 
The input video is rendered from random distance to the origin ($r \in [2, 5]$) with a paired field of view to cover the entire object.
}

\centering
\setlength{\tabcolsep}{6pt}
\footnotesize


\begin{tabular}{llcccccccc}
\toprule
 & &
\multicolumn{4}{c}{Conditioning View} & 
\multicolumn{4}{c}{Novel View} \\
\cmidrule(lr){3-6} \cmidrule(lr){7-10}
& Method & PSNR$\uparrow$ & LPIPS$\downarrow$ & FVD$\downarrow$ & CLIP$\uparrow$
& PSNR$\uparrow$ & LPIPS$\downarrow$ & FVD$\downarrow$ & CLIP$\uparrow$
\\
\midrule

\multirow{7}{*}{\rotatebox{90}{\shortstack{Obj \\ \texttt{random}}}}
& L4GM~\cite{ren2024l4gm} & 22.59 ± 4.74 & 0.074 ± 0.058 & \cellcolor{tabfirst}184 ± 180 & \cellcolor{tabthird}0.95 ± 0.04 & 18.30 ± 4.85 & 0.146 ± 0.072 & 498 ± 261 & 0.90 ± 0.04 \\
& GVF~\cite{zhang2025gaussian} & 18.21 ± 4.90 & 0.117 ± 0.053 & 571 ± 390 & 0.89 ± 0.05 & 16.41 ± 4.73 & 0.157 ± 0.064 & 663 ± 327 & 0.88 ± 0.04 \\
& Ours (Euler, 10) & 22.19 ± 5.90 & 0.076 ± 0.045 & 342 ± 317 & 0.94 ± 0.04 & 18.95 ± 6.23 & 0.130 ± 0.067 & 493 ± 302 & 0.91 ± 0.04 \\
& Ours (Heun, 10) & 23.20 ± 5.89 & 0.065 ± 0.040 & 276 ± 307 & 0.95 ± 0.03 & 19.56 ± 6.05 & 0.120 ± 0.060 & 421 ± 295 & \cellcolor{tabthird}0.93 ± 0.03 \\
& Ours (Euler, 20) & \cellcolor{tabthird}23.67 ± 5.82 & \cellcolor{tabthird}0.060 ± 0.035 & 255 ± 338 & 0.95 ± 0.03 & \cellcolor{tabthird}20.18 ± 6.09 & \cellcolor{tabthird}0.110 ± 0.052 & \cellcolor{tabthird}396 ± 301 & 0.93 ± 0.03 \\
& Ours (Heun, 20) & \cellcolor{tabsecond}23.91 ± 5.79 & \cellcolor{tabsecond}0.057 ± 0.033 & \cellcolor{tabsecond}233 ± 338 & \cellcolor{tabfirst}0.96 ± 0.03 & \cellcolor{tabsecond}20.40 ± 6.01 & \cellcolor{tabsecond}0.106 ± 0.051 & \cellcolor{tabfirst}365 ± 289 & \cellcolor{tabfirst}0.94 ± 0.03 \\
& Ours (Heun, 100) & \cellcolor{tabfirst}24.04 ± 5.76 & \cellcolor{tabfirst}0.056 ± 0.032 & \cellcolor{tabthird}245 ± 378 & \cellcolor{tabsecond}0.96 ± 0.03 & \cellcolor{tabfirst}20.62 ± 6.04 & \cellcolor{tabfirst}0.104 ± 0.052 & \cellcolor{tabsecond}373 ± 307 & \cellcolor{tabsecond}0.94 ± 0.03 \\

\addlinespace[4pt]
\midrule
\addlinespace[4pt]

\multirow{7}{*}{\rotatebox{90}{\shortstack{Obj \\ $r=3$}}}

& L4GM~\cite{ren2024l4gm} & \cellcolor{tabfirst}29.59 ± 5.12 & \cellcolor{tabfirst}0.043 ± 0.048 & \cellcolor{tabfirst}117 ± 172 & \cellcolor{tabfirst}0.97 ± 0.04 & 19.56 ± 4.98 & 0.135 ± 0.068 & 455 ± 240 & 0.90 ± 0.04 \\
& GVF~\cite{zhang2025gaussian} & 17.78 ± 5.60 & 0.134 ± 0.065 & 605 ± 393 & 0.89 ± 0.06 & 16.66 ± 5.25 & 0.159 ± 0.068 & 670 ± 341 & 0.88 ± 0.05 \\
& Ours (Euler, 10) & 21.55 ± 6.19 & 0.089 ± 0.058 & 369 ± 426 & 0.94 ± 0.04 & 18.95 ± 6.34 & 0.134 ± 0.072 & 487 ± 319 & 0.91 ± 0.04 \\
& Ours (Heun, 10) & 22.26 ± 6.06 & 0.080 ± 0.050 & 306 ± 392 & 0.95 ± 0.03 & 19.42 ± 6.09 & 0.125 ± 0.062 & 418 ± 290 & \cellcolor{tabthird}0.93 ± 0.03 \\
& Ours (Euler, 20) & 22.81 ± 5.96 & 0.073 ± 0.044 & 286 ± 370 & 0.95 ± 0.03 & \cellcolor{tabthird}20.04 ± 6.05 & \cellcolor{tabthird}0.115 ± 0.056 & \cellcolor{tabthird}380 ± 295 & 0.93 ± 0.03 \\
& Ours (Heun, 20) & \cellcolor{tabthird}22.89 ± 5.86 & \cellcolor{tabthird}0.071 ± 0.042 & \cellcolor{tabthird}265 ± 397 & \cellcolor{tabsecond}0.96 ± 0.03 & \cellcolor{tabsecond}20.07 ± 5.91 & \cellcolor{tabsecond}0.113 ± 0.055 & \cellcolor{tabsecond}370 ± 298 & \cellcolor{tabfirst}0.94 ± 0.03 \\
& Ours (Heun, 100) & \cellcolor{tabsecond}23.04 ± 5.88 & \cellcolor{tabsecond}0.069 ± 0.040 & \cellcolor{tabsecond}243 ± 299 & \cellcolor{tabthird}0.96 ± 0.03 & \cellcolor{tabfirst}20.24 ± 5.85 & \cellcolor{tabfirst}0.109 ± 0.052 & \cellcolor{tabfirst}353 ± 279 & \cellcolor{tabsecond}0.94 ± 0.03 \\

\addlinespace[4pt]
\midrule
\addlinespace[4pt]

\multirow{7}{*}{\rotatebox{90}{\shortstack{Obj \\ $r=4$}}}

& L4GM~\cite{ren2024l4gm} & 22.47 ± 5.01 & 0.072 ± 0.051 & \cellcolor{tabfirst}180 ± 140 & \cellcolor{tabfirst}0.96 ± 0.04 & 18.38 ± 5.00 & 0.143 ± 0.067 & 505 ± 261 & 0.90 ± 0.04 \\
& GVF~\cite{zhang2025gaussian} & 18.29 ± 5.07 & 0.118 ± 0.058 & 607 ± 417 & 0.89 ± 0.05 & 16.49 ± 4.89 & 0.159 ± 0.069 & 668 ± 330 & 0.88 ± 0.04 \\
& Ours (Euler, 10) & 22.37 ± 5.77 & 0.073 ± 0.042 & 322 ± 337 & 0.94 ± 0.05 & 19.13 ± 6.12 & 0.129 ± 0.068 & 474 ± 303 & 0.91 ± 0.04 \\
& Ours (Heun, 10) & 23.48 ± 5.72 & 0.062 ± 0.036 & 272 ± 339 & 0.95 ± 0.03 & 19.72 ± 5.86 & 0.119 ± 0.058 & 404 ± 271 & \cellcolor{tabthird}0.93 ± 0.03 \\
& Ours (Euler, 20) & \cellcolor{tabthird}23.81 ± 5.78 & \cellcolor{tabthird}0.058 ± 0.033 & 258 ± 318 & 0.95 ± 0.04 & \cellcolor{tabthird}20.25 ± 5.87 & \cellcolor{tabthird}0.110 ± 0.054 & \cellcolor{tabthird}375 ± 291 & 0.93 ± 0.03 \\
& Ours (Heun, 20) & \cellcolor{tabsecond}24.05 ± 5.79 & \cellcolor{tabsecond}0.056 ± 0.033 & \cellcolor{tabthird}251 ± 357 & \cellcolor{tabsecond}0.96 ± 0.03 & \cellcolor{tabsecond}20.36 ± 5.77 & \cellcolor{tabsecond}0.108 ± 0.053 & \cellcolor{tabsecond}362 ± 284 & \cellcolor{tabfirst}0.94 ± 0.03 \\
& Ours (Heun, 100) & \cellcolor{tabfirst}24.31 ± 5.71 & \cellcolor{tabfirst}0.053 ± 0.031 & \cellcolor{tabsecond}237 ± 327 & \cellcolor{tabthird}0.96 ± 0.03 & \cellcolor{tabfirst}20.58 ± 5.78 & \cellcolor{tabfirst}0.104 ± 0.052 & \cellcolor{tabfirst}348 ± 275 & \cellcolor{tabsecond}0.94 ± 0.03 \\

\addlinespace[4pt]
\midrule
\addlinespace[4pt]

\multirow{7}{*}{\rotatebox{90}{\shortstack{C4D \\ \texttt{random}}}}

& L4GM~\cite{ren2024l4gm} & 21.87 ± 5.46 & 0.080 ± 0.053 & \cellcolor{tabfirst}118 ± 46 & \cellcolor{tabfirst}0.98 ± 0.01 & 17.71 ± 2.39 & 0.147 ± 0.060 & \cellcolor{tabthird}432 ± 202 & 0.93 ± 0.04 \\
& GVF~\cite{zhang2025gaussian} & 17.76 ± 4.91 & 0.122 ± 0.076 & 390 ± 160 & 0.93 ± 0.04 & 15.80 ± 3.49 & 0.163 ± 0.078 & 584 ± 216 & 0.92 ± 0.03 \\
& Ours (Euler, 10) & 20.79 ± 6.43 & 0.098 ± 0.077 & 322 ± 193 & 0.93 ± 0.08 & 17.20 ± 4.49 & 0.148 ± 0.079 & 643 ± 228 & 0.88 ± 0.06 \\
& Ours (Heun, 10) & 22.65 ± 4.85 & \cellcolor{tabthird}0.078 ± 0.053 & 190 ± 112 & 0.97 ± 0.01 & 18.13 ± 4.46 & 0.137 ± 0.081 & 549 ± 247 & 0.93 ± 0.03 \\
& Ours (Euler, 20) & \cellcolor{tabthird}22.81 ± 4.91 & 0.079 ± 0.055 & 255 ± 204 & 0.97 ± 0.02 & \cellcolor{tabthird}18.96 ± 3.38 & \cellcolor{tabthird}0.125 ± 0.061 & 490 ± 222 & \cellcolor{tabsecond}0.94 ± 0.03 \\
& Ours (Heun, 20) & \cellcolor{tabfirst}23.32 ± 4.39 & \cellcolor{tabfirst}0.070 ± 0.045 & \cellcolor{tabsecond}169 ± 105 & \cellcolor{tabsecond}0.98 ± 0.01 & \cellcolor{tabfirst}19.31 ± 2.96 & \cellcolor{tabfirst}0.115 ± 0.053 & \cellcolor{tabfirst}387 ± 164 & \cellcolor{tabfirst}0.95 ± 0.02 \\
& Ours (Heun, 100) & \cellcolor{tabsecond}22.95 ± 4.43 & \cellcolor{tabsecond}0.072 ± 0.046 & \cellcolor{tabthird}188 ± 143 & \cellcolor{tabthird}0.98 ± 0.01 & \cellcolor{tabsecond}18.98 ± 3.18 & \cellcolor{tabsecond}0.120 ± 0.057 & \cellcolor{tabsecond}431 ± 198 & \cellcolor{tabthird}0.94 ± 0.02 \\

\addlinespace[4pt]
\midrule
\addlinespace[4pt]

\multirow{7}{*}{\rotatebox{90}{\shortstack{C4D \\ $r=3$}}}

& L4GM~\cite{ren2024l4gm} & \cellcolor{tabfirst}30.14 ± 2.87 & \cellcolor{tabfirst}0.039 ± 0.024 & \cellcolor{tabfirst}77 ± 45 & \cellcolor{tabfirst}0.99 ± 0.01 & 18.33 ± 2.75 & 0.135 ± 0.047 & \cellcolor{tabthird}437 ± 211 & 0.93 ± 0.03 \\
& GVF~\cite{zhang2025gaussian} & 17.57 ± 3.91 & 0.129 ± 0.084 & 446 ± 222 & 0.95 ± 0.02 & 16.28 ± 3.04 & 0.147 ± 0.072 & 576 ± 350 & 0.93 ± 0.03 \\
& Ours (Euler, 10) & 18.83 ± 5.01 & 0.131 ± 0.103 & 482 ± 518 & 0.95 ± 0.03 & 16.36 ± 3.97 & 0.165 ± 0.097 & 715 ± 371 & 0.89 ± 0.02 \\
& Ours (Heun, 10) & 20.06 ± 4.37 & 0.112 ± 0.083 & 257 ± 165 & 0.96 ± 0.02 & 17.07 ± 3.63 & 0.149 ± 0.087 & 594 ± 339 & 0.93 ± 0.02 \\
& Ours (Euler, 20) & \cellcolor{tabthird}22.21 ± 3.54 & 0.084 ± 0.054 & 258 ± 140 & \cellcolor{tabsecond}0.97 ± 0.02 & \cellcolor{tabthird}19.43 ± 3.18 & \cellcolor{tabthird}0.115 ± 0.062 & 437 ± 255 & \cellcolor{tabthird}0.94 ± 0.03 \\
& Ours (Heun, 20) & 22.20 ± 3.09 & \cellcolor{tabthird}0.080 ± 0.046 & \cellcolor{tabthird}215 ± 134 & \cellcolor{tabthird}0.97 ± 0.01 & \cellcolor{tabsecond}19.65 ± 2.89 & \cellcolor{tabsecond}0.109 ± 0.054 & \cellcolor{tabfirst}373 ± 174 & \cellcolor{tabfirst}0.95 ± 0.02 \\
& Ours (Heun, 100) & \cellcolor{tabsecond}22.33 ± 2.90 & \cellcolor{tabsecond}0.078 ± 0.044 & \cellcolor{tabsecond}208 ± 135 & 0.97 ± 0.02 & \cellcolor{tabfirst}19.75 ± 2.87 & \cellcolor{tabfirst}0.108 ± 0.052 & \cellcolor{tabsecond}383 ± 216 & \cellcolor{tabsecond}0.95 ± 0.02 \\

\addlinespace[4pt]
\midrule
\addlinespace[4pt]

\multirow{7}{*}{\rotatebox{90}{\shortstack{C4D \\ $r=4$}}}

& L4GM~\cite{ren2024l4gm} & 21.62 ± 4.79 & 0.075 ± 0.047 & \cellcolor{tabfirst}123 ± 51 & \cellcolor{tabfirst}0.98 ± 0.01 & 17.10 ± 2.85 & 0.145 ± 0.055 & 475 ± 172 & 0.92 ± 0.03 \\
& GVF~\cite{zhang2025gaussian} & 17.74 ± 3.52 & 0.112 ± 0.070 & 410 ± 207 & 0.93 ± 0.05 & 16.25 ± 3.24 & 0.145 ± 0.077 & 511 ± 265 & 0.93 ± 0.04 \\
& Ours (Euler, 10) & 20.94 ± 4.92 & 0.090 ± 0.057 & 406 ± 183 & 0.95 ± 0.03 & 16.56 ± 4.00 & 0.159 ± 0.088 & 663 ± 300 & 0.90 ± 0.03 \\
& Ours (Heun, 10) & 22.67 ± 4.00 & 0.075 ± 0.053 & 261 ± 173 & \cellcolor{tabsecond}0.97 ± 0.01 & 17.11 ± 3.80 & 0.147 ± 0.081 & 600 ± 274 & 0.93 ± 0.03 \\
& Ours (Euler, 20) & \cellcolor{tabsecond}24.53 ± 3.42 & \cellcolor{tabthird}0.062 ± 0.038 & \cellcolor{tabthird}191 ± 108 & \cellcolor{tabthird}0.97 ± 0.02 & \cellcolor{tabthird}19.92 ± 3.58 & \cellcolor{tabthird}0.114 ± 0.065 & \cellcolor{tabthird}435 ± 220 & \cellcolor{tabthird}0.94 ± 0.03 \\
& Ours (Heun, 20) & \cellcolor{tabfirst}24.58 ± 3.02 & \cellcolor{tabfirst}0.058 ± 0.034 & 209 ± 168 & 0.97 ± 0.01 & \cellcolor{tabfirst}20.58 ± 3.15 & \cellcolor{tabfirst}0.101 ± 0.054 & \cellcolor{tabfirst}346 ± 148 & \cellcolor{tabfirst}0.96 ± 0.02 \\
& Ours (Heun, 100) & \cellcolor{tabthird}24.25 ± 3.17 & \cellcolor{tabsecond}0.060 ± 0.036 & \cellcolor{tabsecond}171 ± 76 & 0.97 ± 0.02 & \cellcolor{tabsecond}20.53 ± 3.06 & \cellcolor{tabsecond}0.102 ± 0.052 & \cellcolor{tabsecond}367 ± 151 & \cellcolor{tabsecond}0.95 ± 0.02 \\
\bottomrule
\end{tabular}
\label{tab:supp_generation}
\end{table*}

\begin{table}[t]
\centering
\setlength{\tabcolsep}{3pt}
\resizebox{\linewidth}{!}{
\begin{tabular}{lcccc}
\toprule
Method &
$L_{2,\text{all}}^{3D}\,\downarrow$ &
$\text{APD}_\text{all}^\text{3D}\,\uparrow$ &
$\text{AJ}^\text{3D}\,\uparrow$ &
$\text{OA}\,\uparrow$ \\
\midrule
SpatialTrackerV2 &
\cellcolor{tabsecond}0.0704 &
37.52 &
28.83 &
\cellcolor{tabsecond}0.8582 \\
Cotracker3, GT depth &
0.1941 &
\cellcolor{tabsecond}49.44 &
\cellcolor{tabsecond}40.04 &
0.8182 \\
Ours &
\cellcolor{tabfirst}0.0406 &
\cellcolor{tabfirst}49.72 &
\cellcolor{tabfirst}40.24 &
\cellcolor{tabfirst}0.9388 \\
\bottomrule
\end{tabular}
}
\vspace{-0.5em}
\caption{\textbf{3D tracking metrics on real data from TAPVid-Panoptic~\cite{koppula2024tapvid3d}}. Our method achieves the best overall performance across all reported metrics.}
\label{tab:real_tracking}
\vspace{-1em}
\end{table}

\begin{table}[t]
\centering
\setlength{\tabcolsep}{3pt}
\resizebox{\linewidth}{!}{
\begin{tabular}{ccccc}
\hline
\# Latents & Runtime & Chamfer $\downarrow$ & LPIPS $\downarrow$ & PSNR $\uparrow$ \\
\hline
$4096$  & 0.19s & 4.33 & 0.032 & 33.1 \\
$8192$  & 0.38s & \cellcolor{tabthird}3.88 & \cellcolor{tabthird}0.029 & \cellcolor{tabthird}33.9 \\
$16384$ & 1.11s & \cellcolor{tabfirst}3.69 & \cellcolor{tabfirst}0.026 & \cellcolor{tabfirst}34.8 \\
\hline
\end{tabular}
}
\vspace{-0.5em}
\caption{\textbf{Token size ablation (Ours-S).} Increasing the number of latent tokens improves geometric and perceptual quality, measured by Chamfer, LPIPS, and PSNR, at the cost of higher runtime. Chamfer is scaled by $10^4$.}
\label{tab:token_ablation}
\vspace{-1em}
\end{table}

\begin{table*}[t]
\centering
\setlength{\tabcolsep}{3pt}
\resizebox{\linewidth}{!}{
\begin{tabular}{lcccccccc}
\toprule
Dataset & Chamfer $\downarrow$ & Chamfer oracle $\downarrow$ & Chamfer change & LPIPS $\downarrow$ & PSNR $\uparrow$ & SSIM $\uparrow$ & CVVDP $\uparrow$ & FVD $\downarrow$ \\
\midrule
PokeFlex (GT vox.)   & 4.64 & 4.07 & 14\% & 0.0493 & 34.58 & 0.964 & 8.92  & 59.1 \\
PokeFlex (dec. vox.)  & 4.64 & 4.07 & 14\% & 0.0493 & 34.42 & 0.963 & 8.90  & 62.3 \\
\midrule
Objaverse (GT vox.)  & 3.64 & 3.11 & 17\% & 0.0210 & 35.39 & 0.984 & 8.91 & 49.0 \\
Objaverse (dec. vox. ) & 3.64 & 3.11 & 17\% & 0.0220 & 35.11 & 0.983 & 8.88 & 50.7 \\
\bottomrule
\end{tabular}
}
\vspace{-0.5em}
\caption{\textbf{Reconstruction performance across datasets.} We evaluate our method (trained on Objaverse) on the PokeFlex~\cite{obrist2025pokeflex} dataset, to show the versatility of the learnt tokens. GVF cannot be applied due to missing temporal correspondences.}
\label{tab:pokeflex_result}
\vspace{-1em}
\end{table*}

\subsection{Real-world results}
%
We show experiments on real-world data and are encouraged by the results.
For video-to-4D (\cref{fig:real_world}), our method qualitatively outperforms baselines including L4GM, which relies on ImageDream, a model partially trained on real data.
For 3D tracking, we evaluate on segmented videos from TAPVid-3D Panoptic Studio~\cite{koppula2024tapvid3d}. Our tracking method is competitive with SOTA methods, which train on real data (see \cref{tab:real_tracking}).
%
%
We evaluate 4D reconstruction on the PokeFlex dataset~\cite{obrist2025pokeflex}, derived from real RGBD scans.
%
Our method achieves a PSNR of $34.4$dB, only \textbf{2\%} degradation compared to Objaverse, suggesting that the sim-to-real gap may not be significant (see \cref{tab:pokeflex_result}). 
Notably, the deformation-based GVF~\cite{zhang2025gaussian} cannot be applied as it needs temporal correspondences that are unavailable in PokeFlex, demonstrating Velox's improved accessibility.

Broadly, our overarching goal is to learn a representation of ideal dynamic objects, free of holes and other artifacts, so downstream models can directly reason about object dynamics. 
Today, such data is mainly available in synthetic form~\cite{deitke2022objaverse}. 
Nonetheless, as shown by our results, training on such data already enables useful applications like 4D object generation.
Incorporating real training data is an important future direction.

\subsection{Token size ablation} 
We conduct a latent size ablation (see \cref{tab:token_ablation}). 
As expected, the performance improves with increasing the latent representation size.
Among these, using $8192$ latents offers a good balance between reconstruction quality and runtime, and it also matches latent sizes of recent 3D/4D representations~\cite{chang2026lito, zhang2025gaussian} for fair comparison.

\subsection{Video-to-4D}
Please find additional video-to-4D results in the supplementary website and the full metrics in~\ref{tab:supp_generation}. 

We evaluate the same model on various input camera settings and two datasets, Objaverse (Obj) and Consistent4D~\cite{jiang2024consistentd} (C4D).
Across all settings, our model provides consistently strong video-to-4D generation, with small variations depending on the input camera pose setting.
Under \texttt{random} camera positions, where the distance to the origin ranges from 2 to 5 units, our model produces stronger performance in both input view alignment and novel view quality. 
%
Note that our model does not take camera parameters (\eg, pinhole position, field of view, ray maps, etc) as input, and the model needs to infer this information from the input image if needed.
%
%

When the input camera is at a fixed location at $r = 3$, 
L4GM achieves better input view alignment than ours.
This is due to its pixel-aligned design, which benefits input-view alignment, and the camera setting matches their training setup.
Additionally, L4GM relies on an azimuth search that optimizes the metrics between the rendered and the given input views.
On the other hand, our model does not use any search/optimization during inference.
Nonetheless, our method provides the best novel-view performance, indicating a robust ability to estimate 4D shapes and appearance.

Across \texttt{random}, $r=3$ and $r=4$ settings, our approach offers a significant improvement for novel views, indicating higher quality of the overall generated 4D objects.
This is due to the fact that we model the distributions of the full 4D representations, whereas L4GM models multiview image distributions, and GVF models distributions of individual Gaussian movements across frames, which is a difficult task when the number of Gaussians is large.

Additionally, the table also indicates the various performance effects due to both the number of ODE steps and the types of integration steps (Heun vs Euler). The paper reports values of Heun with 100 steps, however we can see that both Heun and Euler integration modes with a lower number of steps already outperform baselines on novel view perceptual metrics. 
\subsection{3D tracking}

Please find additional tracking results in the supplemental webpage and~\cref{tab:supp_tracking}. 
We include results of raw, unfiltered, SpatialTrackerv2~\cite{xiao2025spatialtrackerv2} and Cotracker3~\cite{karaev2024cotracker3}, where we do not remove tracks outside $[-1, 1]^3$ from the calculation, as we do in the main paper. 
As we can see, the filtering significantly improves the baselines, by removing floating-pixel artifacts these methods often exhibit.
In the table, we include additional 2D tracking evaluation results.
As can be seen, our method outperforms the baselines across all 2D metrics as well. 
We again attribute this to the fact that we purely track in our latent space, which jointly models both 4D geometry and appearance. 
This is useful when objects have only simple textures.

\subsection{Cloth simulation}
Please see~\cref{fig:supp_cloth} for additional cloth simulation results and~\cref{tab:cloth_to_4d} for the quantitative cloth simulation results. 
Similar to the video-to-4D evaluation, we measure standard image reconstruction metrics, including PSNR, LPIPS, SSIM, and CLIP. 
Additionally, we evaluate physics-based metrics such as the Root Mean Square Error (RMSE) on the position of the center of mass along the gravity direction.
We first compute the center of mass of the object from the output point cloud and then calculate its position over time. 
The reported errors are measured against the reference physics simulation results from the full 3D clothes.

While the generated cloth may have a different full 3D shape than the reference, as the conditioning image often does not see the full cloth, the metrics indicate that our method closely captures how the cloth would fall and impact the ground. 
This demonstrates that our approach achieves a degree of physical realism and highlights the versatility of our tokens, even for physics-based simulations.

\begin{table*}[t]
\captionof{table}{\textbf{Cloth simulation evaluation} We evaluate the cloth simulation via image-to-4D for the generative metrics. We skip FVD, as the data distribution is far from what the FVD feature model (InceptionI3d) was trained on. 
}
\vspace{-3mm}
\centering
\setlength{\tabcolsep}{6pt}
\footnotesize


\begin{tabular}{lccccccc}
\toprule
 & 
\multicolumn{3}{c}{Conditioning View} & 
\multicolumn{3}{c}{Novel View} \\
\cmidrule(lr){2-4} \cmidrule(lr){5-7} & PSNR$\uparrow$ & LPIPS$\downarrow$ & CLIP$\uparrow$
& PSNR$\uparrow$ & LPIPS$\downarrow$ &  CLIP$\uparrow$ & RMSE Position$\downarrow$
\\
\midrule
\multirow{3}{*}{\rotatebox{90}{\shortstack{}}}
 & 24.25 ± 2.50 & 0.031 ± 0.012 & 0.944±0.019 & 22.92 ± 2.58 & 0.041 ± 0.016 & 0.930±0.021 & 0.012 ± 0.008 \\
\bottomrule
\end{tabular}
\label{tab:cloth_to_4d}
\end{table*}

\section{Additional Related Work}
\label{sec: supp additional related work}
For a summary of the methods on 4D representation learning, please see Table~\ref{tab:related_work}. 
Below we discuss work related to our application areas of 3D tracking and 4D generation.

\subsection{3D tracking}

While 2D point tracking has been extensively studied~\cite{karaev2024cotracker,sand2008particle,karaev2024cotracker3,harley2022particle,le2024dense}, extending these ideas to 3D requires representations that capture spatial consistency across time.
OmniMotion~\cite{wang2023tracking} demonstrates accurate 3D point tracking by introducing a canonical 3D volume with bijective mappings to each frame’s local volume. However, its reliance on test-time optimization limits scalability.
Subsequent works~\cite{xiao2024spatialtracker,wang2024scenetracker,ngo2024delta} avoid test-time optimization through feed-forward designs that propagate point or trajectory features over time iteratively. 
%
Recent approaches often track directly in world coordinate space~\cite{zhang2025tapip3d, jin2024stereo4d,feng2025startrack,wang2024dust3r,xiao2024spatialtracker}, enabling stable long-range trajectories. 
SpatialTrackerV2~\cite{xiao2024spatialtracker} further enhances performance by integrating dual 2D and 3D tracking heads with cross-attention for feature exchange. 

Our representation is learned without tracking-specific supervision or temporal correspondences. 
Nevertheless, it supports accurate 3D tracking through a simple feed-forward model applied over the learned 4D representation.

\subsection{4D generation}

A large body of recent work addresses video- or image-to-4D generation~\cite{jiang2024consistentd,bah20244dfy,ren2023dreamgaussian4d,singer2023text4d,bah2024tc4d,gao2024gaussianflow,ling2024align,yin20234dgen,zhao2023animate124,zheng2024unified, zeng2024stag4d,wu2024cat4d,xie2024sv4d,yao2024sv4d2,pan2024efficient4d,liang2024diffusion4d}. Most methods rely on pretrained multiview generative priors~\cite{liu2023zero,shi2023zero123plus,shi2023MVDream,wang2023imagedream}, producing multiview images or videos that are subsequently fed into either optimization-based or feed-forward 4D reconstruction pipelines. 
Because these generated views are often imperfectly aligned, resulting 4D reconstructions frequently inherit artifacts or temporal inconsistencies. 
Similar challenges arise in a recent work L4GM~\cite{ren2024l4gm}, which directly predicts per-frame, pixel-aligned 3D Gaussians.  

Beyond view-synthesis–driven approaches, recent work has begun learning generative models over 4D object representations. 
GVF~\cite{zhang2025gaussian} generates 3D Gaussian splats for the first video frame and predicts a deformation field that evolves them over time; as a result, unobserved regions in the initial frame may be inconsistent with later observations. 
ShapeGen4D~\cite{yenphraphai2025shapegen} jointly generates 3D representations for all frames but models geometry only and requires tracking information during training. 
Both methods additionally rely on inference-time optimization to align generated objects to input views.
In contrast, our approach directly generates a coherent 4D representation that aligns with the input view, without relying on tracking information during training.

For deformable object simulation, several works use mesh-based representations and graph neural networks~\cite{zhang2024adaptigraph, huang2022mesh, helearning, shi2024robocraft, shi2023robocook, ai2024robopack} to model object manipulation, while more recent efforts explore generative approaches~\cite{tian2025uniclothdiff}. 
These methods typically address control or manipulation under targeted external forces. 
Our setting differs: we focus on simulating cloth dynamics conditioned solely on an initial image, framing the task as an image-to-4D prediction problem rather than a control or manipulation problem.

\section{Author contributions}
\label{sec:author_contrib}
All authors conceived the method and its applications. 
Anagh Malik and Jen-Hao Rick Chang implemented the method and experimental codebase, including dataset rendering and baselines. 
Anagh Malik conducted the experiments during the project, and Jen-Hao Rick Chang ran the final experiments presented in the paper. 
Dorian Chan contributed to the baselines and assisted with the final tracking experiments. 
Anagh Malik prepared the draft of the manuscript and all figures, which were subsequently refined by Dorian Chan. 
David B. Lindell, Oncel Tuzel, and Jen-Hao Rick Chang supervised the project and helped revise and polish the manuscript. 
All authors contributed to discussions, experiment design, and the paper.

\end{document}